\documentclass{article}

\usepackage[final,nonatbib]{neurips_2022}

\usepackage[utf8]{inputenc} % allow utf-8 input
\usepackage[T1]{fontenc}    % use 8-bit T1 fonts
\usepackage{hyperref}       % hyperlinks
\usepackage{url}            % simple URL typesetting
\usepackage{booktabs}       % professional-quality tables
\usepackage{amsfonts}       % blackboard math symbols
\usepackage{nicefrac}       % compact symbols for 1/2, etc.
\usepackage{microtype}      % microtypography
\usepackage{xcolor}         % colors
\usepackage{amsmath}
\usepackage{amsthm}
\usepackage{amsfonts}
\usepackage{graphicx}
\usepackage{subcaption}
\usepackage{multicol}
\usepackage{multirow}
\usepackage{algorithm}
\usepackage{algorithmic}
\usepackage{wrapfig}

\newtheorem{lemma}{Lemma}

\renewcommand{\paragraph}[1]{\noindent\vspace{-0.0em}\textbf{{#1}}}

\title{Optimal Brain Compression: A Framework for Accurate Post-Training Quantization and Pruning}

\author{%
    Elias Frantar \thanks{Corresponding author.} \\
    IST Austria \\
    \texttt{elias.frantar@ist.ac.at}
    \And 
    Sidak Pal Singh \\
    ETH Zurich \\
    \texttt{sidak.singh@inf.ethz.ch}
    \And
    Dan Alistarh \\
    IST Austria \& Neural Magic \\
    \texttt{dan.alistarh@ist.ac.at}
}

\begin{document}

\maketitle

\begin{abstract}
We consider the problem of model compression for deep neural networks (DNNs) in the challenging one-shot/post-training setting, in which we are given an accurate trained model, and must compress it without any retraining, based only on a small amount of calibration input data. This problem has become popular in view of the emerging software and hardware support for executing models compressed via pruning and/or quantization with speedup, and well-performing solutions have been proposed independently for both compression approaches.
In this paper, we introduce a new compression framework which covers both weight pruning and quantization in a unified setting, is time- and space-efficient, and considerably improves upon the practical performance of existing post-training methods. At the technical level, our approach is based on an exact and efficient realization of the classical Optimal Brain Surgeon (OBS) framework of [LeCun, Denker, and Solla, 1990] extended to also cover weight quantization at the scale of modern DNNs. From the practical perspective, our experimental results show that it can improve significantly upon the compression-accuracy trade-offs of existing post-training methods, and that it can enable the accurate \emph{compound} application of both pruning and quantization in a post-training setting.
\end{abstract}

\section{Introduction}

The impressive recent progress of deep learning for solving challenging tasks across several domains has been accompanied by a significant increase in  parameter counts and computational costs for executing such models. A natural consequence has been a growing effort to reduce such costs via 
\emph{model compression}, and 
the two most popular approaches for model compression are \emph{pruning}---removing neural network weights by setting them to zero---and \emph{quantization}, reducing the precision at which neural network weights and activations are stored and manipulated. Hundreds of such pruning and quantization approaches have been proposed and analyzed~\cite{hoefler2021sparsity, gholami2021survey}, with the general goal of obtaining efficient variants of deep neural nets (DNNs) which would preserve accuracy while maximizing compression. Despite impressive progress, compression is still a laborious process: the pruning and quantization stages are often done independently, and recovering model accuracy after compression often requires partial or even full retraining of the compressed model. 

An alternative but challenging scenario is the \emph{post-training compression} setup~\cite{nagel2020up, li2021brecq, hubara2021accurate, liang2021pruning}, in which we are given a trained but uncompressed model, together with a small amount of \emph{calibration data}, and must produce an accurate compressed model in \emph{one shot}, i.e. a single compression step, without retraining, and with limited computational costs. This is motivated by practical scenarios such as the MLPerf Inference Benchmark~\cite{reddi2020mlperf}, 
and is the setting we focus on in this paper.  

Compression via \emph{weight pruning} started with  seminal work by LeCun et al.~\cite{lecun1990optimal}, complemented by Hassibi and Stork~\cite{hassibi1993optimal}, who proposed a mathematical framework called the \emph{Optimal Brain Surgeon (OBS)}, for choosing the set of weights to remove from a trained neural network, by leveraging second-order information. (We describe their approach in Section~\ref{sec:obs}.) 
Recent advances, e.g.~\cite{2017-dong, wang2019eigendamage, singh2020woodfisher, frantar2021m} showed that OBS can lead to state-of-the-art compression at DNN scale, by introducing numerical methods which can approximate the second-order information needed by OBS at the massive parameter counts of modern models. 
However, these approaches do not apply to the \emph{post-training} setting, as they require gradual pruning, as well as significant retraining,  to recover good accuracy.

An alternative approach, which is standard in the context of \emph{post-training} compression, 
has been to break the compression task into layer-wise sub-problems, identifying a compressed weight approximation for each layer, given a sub-sample of the layer's inputs and outputs based on calibration data. 
This line of work, e.g.~\cite{wang2020towards, nagel2020up, hubara2021accurate}, introduced elegant solvers for the resulting layer-wise weight quantization problem, which achieve state-of-the-art results for  post-training quantization. 
Recently, AdaPrune~\cite{hubara2021accelerated} showed that this approach can also be effective for post-training weight pruning. 

In this context, a natural question is whether existing approaches for pruning and quantization can be \emph{unified} in order to cover both types of compression in the post-training setting, thus making DNN compression simpler and, hopefully, more accurate. 
This question is also of practical importance, since both GPU and CPU platforms now \emph{jointly} support sparse and quantized formats~\cite{NVIDIASparse, deepsparse}, and, as we illustrate experimentally, the resulting models could be executed with compound speedups. 

\paragraph{Contribution.} 
In this paper, we provide a mathematical framework for compression via pruning or quantization, which leads to state-of-the-art accuracy-versus-compression trade-offs in the challenging \emph{post-training compression} setup.
Our framework starts from the layer-wise compression problem described above, by which the global compression task, defined either for pruning or quantization, is first split into layer-wise sub-problems, based on the layer behavior on the calibration data. 
Specifically, given a layer $\ell$ defined by  weights $\mathbf{W}_\ell$, and layer inputs $\mathbf{X}_\ell$, we aim to find a compressed version of the weights $\mathbf{\widehat{W}}_\ell$ which minimizes the output difference relative to the uncompressed layer, measured via the squared error between the original and compressed layer, acting on the sample input 
$|| \mathbf{W_\ell} \mathbf{X_\ell} - \mathbf{\widehat{W}_\ell} \mathbf{X_\ell}||_2^2$, under a fixed compression constraint on $\mathbf{\widehat{W}_\ell}$.

Although solving this problem optimally for sparsity or quantization constraints is NP-hard~\cite{blumensath2008iterative, nagel2020up}, it is a key step in all state-of-the-art post-training compression methods, both for pruning~\cite{hubara2021accelerated, frantar2022spdy} and for quantization~\cite{nagel2020up, hubara2021accelerated, li2021brecq}. Once this is solved per layer, a solution to the global problem can be obtained by combining layer-wise solutions, which is handy especially for non-uniform compression, e.g.~\cite{he2018amc, frantar2022spdy}.
Thus, several approximations for this problem have been proposed~\cite{nagel2020up, hubara2021accurate, hubara2021accelerated}.  

We show that there still is significant room for improvement when solving the layer-wise compression problem.
Roughly, our approach is to specialize the OBS framework to the squared error formulation above: in this case, the framework can in theory produce an exact greedy solution, but a direct implementation would have infeasible $\Theta(d^4)$ computational cost, where $d$ is the layer dimension. 
Our main technical contribution is a series of algorithms which reduce this computational cost, \emph{without any approximations}, to $O(d \cdot d_{col}^2)$ where $d_{col}$ is the column dimension of the weight matrix. 
In practice, these improvements are significant enough to allow us to  implement the exact OBS greedy solution, which prunes \emph{one weight at a time}, and updates \textit{all remaining weights} after each step, at the scale of modern DNNs with tens of millions of parameters, within reasonable time, on a single GPU. We provide efficient implementations of our methods at \url{https://github.com/IST-DASLab/OBC}.

In turn, this algorithmic development allows us to apply the OBS approach to \emph{quantization}. 
The resulting algorithm, called the \emph{Optimal Brain Quantizer (OBQ)}, quantizes weights iteratively one-at-a-time, depending on their impact on the loss increase, after which it applies a closed-form update to the remaining unquantized weights,  further reducing the loss. 
This solves the two problems efficiently, and in a unified manner---we call the unified framework the \emph{Optimal Brain Compressor (OBC)}.

\paragraph{Experimental Results.}
We apply OBC to standard tasks and models covering image classification, object detection, and language modelling applications. 
We first show that our framework yields significantly better solutions for the layer-wise compression problem, which 
 leads to higher-accuracy end-to-end compressed models for both pruning and quantization, relative to the corresponding state-of-the-art techniques, often by significant margins.
Second, we show that our pruning and quantization approaches can be compounded, with surprisingly strong results: we obtain a 12$\times$ reduction in theoretical operations with a 2\% accuracy drop for GPU-supported compound compression~\cite{NVIDIASparse}, and a 4$\times$  speedup in \emph{actual runtime} with only $1\%$ accuracy drop for a CPU-based sparsity-aware runtime~\cite{deepsparse}. Together, these results suggest for the first time that  post-training compression can be competitive with full retraining. 

\section{Related Work}

\paragraph{Optimal Brain Surgeon (OBS).}
The classic OBS framework \cite{lecun1990optimal, hassibi1993optimal} was originally applied to networks with hundreds of weights; more recently, methods such as  WoodFisher \cite{singh2020woodfisher} rendered the approach computationally feasible for DNNs by using a block-diagonal Fisher approximation of the Hessian, while follow-up methods introduced more efficient and general algorithms for handling the inverse Fisher matrix \cite{frantar2021m}, or customize this approximation to specific model families~\cite{kurtic2022optimal}. 
 Earlier work called Layer-wise OBS (L-OBS)~\cite{2017-dong} was inspired by the  K-FAC approximation~\cite{2015-martens, grosse2016kroneckerfactored}: L-OBS approximates the OBS framework not for the global objective, but for a quadratic per-layer loss, while also pruning all weights based on a single Hessian computation. At a high level, our approach is similar, in that we apply OBS layer-wise; however, we apply OBS \textit{exactly}, that is, pruning one weight at a time, and exactly recomputing the Hessian after every pruning step. This is made computationally tractable by several new algorithmic ideas, and yields significantly improved results relative to L-OBS.
This prior work on pruning considered settings with {extensive finetuning}. By contrast, we will focus on the post-training setting, where only a small amount of calibration data is available.

\paragraph{Post-Training Quantization.}
This setting has been primarily considered for quantization, and most state-of-the-art methods work by performing layer-wise compression. 
Specifically, BitSplit~\cite{2017-dong} optimizes the quantized weights bit by bit, while AdaRound \cite{nagel2020up} finds a weight rounding policy through gradient based optimization with an annealed penalty term that encourages weights to move towards points on the quantization grid. AdaQuant \cite{hubara2021accurate} relaxes the AdaRound constraint, allowing weights to change during quantization-aware optimization, via  straight-through estimation~\cite{nagel2021white}. BRECQ~\cite{li2021brecq} suggested that accuracy can be improved further by integrating second-order information into the layer-wise losses and by jointly optimizing hand-crafted blocks of related layers. 

A key step of AdaRound, AdaQuant and BRECQ is to quantize layers incrementally, in  \textit{sequential} order, so that errors accumulated in earlier layers can be compensated by weight adjustments in later ones. This significantly improves performance, but reduces flexibility, as the entire process may need to be re-done whenever we wish to change compression parameters of one layer. 
We instead target \textit{independent} compression of each layer, allowing the end model to be simply ``stitched'' together from layer-wise results.
Despite operating independently on each layer, we find that, after correcting basic statistics such as batchnorm, our method performs on par to sequential ones for uniform quantization.

\paragraph{Post-Training Sparsification.}
The layer-wise approach was shown to also be effective for post-training pruning by AdaPrune~\cite{hubara2021accelerated}, which pruned weights to the GPU-supported N:M pattern \cite{zhou2021learning}. AdaPrune first drops parameters according to their magnitude \cite{zhu2017prune} and then reoptimizes the remaining weights to reconstruct the pre-compression calibration set output. This is similar to \cite{he2017channel, evci2018mean} which also perform  layer-wise reoptimization of the remaining weights. 
Follow-up work \cite{frantar2022spdy} noted that the results of AdaPrune can be improved further by performing more frequent  pruning/optimization steps. Our algorithm pushes this idea to the limit, performing \emph{full reoptimization} after every single pruned weight, while remaining computationally tractable. We further use a more sophisticated weight selection metric which incorporates second-order information. Finally, \cite{frantar2022spdy} also introduces \textit{global AdaPrune}, a more expensive global optimization step applied on top of the layer-wise AdaPrune results, which can bring additional accuracy gains. This can also be applied to our pruned models.

\paragraph{Non-Uniform Compression.} An orthogonal practical question is how to compress different layers to maximize accuracy under a given resource constraint, such as latency or energy consumption. Existing methods can be roughly categorized into search-based and solver-based approaches. The former, e.g. AMC \cite{he2018amc} or HAQ \cite{wanghaq}, search for a layer-wise compression policy directly via, for example, reinforcement learning or genetic programming \cite{yang2020automatic}, whereas the latter, e.g. HAWQv3~\cite{yao2021hawq} or AdaQuant \cite{hubara2021accurate}, construct a relaxed version of the overall problem that is then solved exactly. We focus here on solver-based approaches, as they can rapidly adapt to different scenarios when combined with accurate independent layer-wise compression schemes; however, our techniques could be of interest for search-based methods as well. Concretely, we use the problem formulation of AdaQuant~\cite{hubara2021accurate} to which we apply the DP algorithm of SPDY \cite{frantar2022spdy} to achieve fast solving times even with a large number of possible choices per layer.

\section{Problem Definition and Background}
\label{sec:background}

\paragraph{The Layerwise Compression Problem.} 
Following prior work on post-training compression, e.g.~\cite{nagel2020up, hubara2021accurate}, we define the problem as follows. 
Mathematically, we model a layer $\ell$ as a function 
 $f_\ell(X_\ell, W_\ell)$ acting on inputs $X_\ell$, parametrized by weights $W_\ell$. 
 The goal of layer-wise compression is to find a ``compressed'' version of $W_\ell$ that performs as similarly as possible to the original weights. More formally, the compressed weights $\widehat{W}_\ell$ should minimize the expected layer output change as measured by some loss $\mathcal{L}$ while at the same time satisfying a generic compression constraint, which we denote by $\mathcal{C}(\widehat{W}_\ell) > C$, which will be customized depending on the compression type:
\begin{equation}
    \text{argmin}_{\widehat{W}_\ell} \quad \mathbb{E}_{X_\ell} \, \mathcal{L}(f_\ell(X_\ell, W_\ell), f_\ell(X_\ell, \widehat{W}_\ell)) \quad \text{subject to} \quad \mathcal{C}(\widehat{W}_\ell) > C.
\end{equation}
The expectation over the layer inputs $X_\ell$ is usually approximated by taking the mean over a small set of $N$ input samples. This low-data setting is one of the primary applications of layer-wise compression. Further, most works \cite{wang2020towards, nagel2020up, hubara2021accurate} focus on compressing linear and convolutional layers, which can be unfolded into linear ones, as these are prevalent in practice, and use the squared loss to measure the approximation error. 
This definition of the loss can be motivated, via a sequence of approximations, from second-order information: please see~\cite{nagel2020up} for a precise derivation. 
Furthermore, this approximation approach has been shown to work well in many applications \cite{nagel2020up, hubara2021accurate, frantar2022spdy}. 

We follow these conventions as well, and work with the specific layer-wise compression problem stated formally below, where the weights $\mathbf{W_\ell}$ are a $d_\text{row} \times d_\text{col}$ matrix (for conv-layers $d_\text{col}$ corresponds to the total number of weights in a single filter), and the input $\mathbf{X_\ell}$ has dimensions $d_\text{col} \times N$.
\begin{equation}
    \label{eq:lwcomp}
    \text{argmin}_{\mathbf{\widehat{W}_\ell}} \quad ||\mathbf{W_\ell} \mathbf{X_\ell} - \mathbf{\widehat{W}_\ell} \mathbf{X_\ell}||_2^2 \quad \text{s.t.} \quad \mathcal{C}(\mathbf{\widehat{W}_\ell}) > C.
\end{equation}

\paragraph{The Optimal Brain Surgeon (OBS) Framework.} \label{sec:obs}
The OBS framework~\cite{lecun1990optimal, hassibi1993optimal} considers the problem of accurately pruning a trained dense neural network. 
It starts from the Taylor approximation at the given point (assumed to have negligible gradient), and provides explicit formulas for the optimal single weight to remove, as well as the optimal update of the remaining weights which would compensate for the removal. 
More precisely, let $\mathbf{H}$ denote the Hessian matrix of the loss at the given (dense) model. 
Then the weight to prune $w_p$ which incurs the minimal increase in loss and the corresponding update of the remaining weights $\boldsymbol{\delta_p}$ can be calculated as follows:
\begin{equation}
    \label{eq:obs}
    w_p = \text{argmin}_{w_p} \, \frac{w_p^2}{[\mathbf{H}^{-1}]_{pp}}, \quad \boldsymbol{\delta_p} = - \frac{w_p}{[\mathbf{H}^{-1}]_{pp}} \cdot \mathbf{H}^{-1}_{:, p},
\end{equation}
\noindent where $[\mathbf{H}^{-1}]_{pp}$ denotes the $p$th diagonal entry of the inverse Hessian, and $\mathbf{H}^{-1}_{:, p}$ is its $p$th column. 

\paragraph{OBS for Layer-Wise Pruning.} We will now instantiate this framework for the layer-wise pruning problem, defined above. 
First, the loss in equation (\ref{eq:lwcomp}) is quadratic and since our starting point is given by the dense weights achieving the minimal loss of 0, the assumptions of the OBS framework are fully met, meaning that its formulas are \emph{exact} for this specific problem formulation.
Thus, iterating the OBS framework to remove one weight at a time would yield an exact \emph{greedy solution} for the layer-wise pruning problem, as it takes the (locally) optimal decision at each step. 
While this greedy approach does not guarantee convergence to a global optimum, such approaches can be very effective for dealing with problem instances that are too large to be handled by exact methods.

\section{An Optimal Greedy Solver for Sparsity}

The obvious challenge is that applying the OBS framework in its true form, i.e. pruning a single weight at a time using the exact formulas in (\ref{eq:obs}), is computationally very demanding. The Hessian $\mathbf{H}$ is a $d \times d$ matrix where $d = d_\text{row} \cdot d_\text{col}$, which is already expensive to store and compute with. 
Additionally, this matrix needs to be updated and inverted at each of the $O(d)$ steps with a computational complexity of $\Theta(d^3)$. Clearly, an $O(d^4)$ total runtime is too inefficient for pruning most layers of modern neural networks, as $d$ is usually $\geq 10^5$ or even $\geq 10^6$ for several layers. However, as we will now show, it is actually possible to reduce the overall costs of this process to $O(d_\text{row} \cdot d_\text{col}^3)$ time and $\Theta(d_\text{col}^2)$ memory, making it efficient enough to prune e.g. all layers of a medium-sized model such as ResNet50 in a bit more than one hour on a single NVIDIA RTX 3090 GPU. We emphasize that the techniques we introduce are exact; unlike prior work~\cite{2017-dong, singh2020woodfisher}, we do not rely on any approximations.

\label{sec:trueobs}
\paragraph{The ExactOBS Algorithm.}
In the following, we introduce our efficient instantiation of the OBS framework, for the layer-wise compression problem, which we call ExactOBS, in step-by-step fashion.
We start by rewriting the matrix squared error in (\ref{eq:lwcomp}) as the sum of the squared errors for each row in the weight matrix. 
As we are always dealing with a fixed layer $\ell$, we drop the subscript $\ell$ to simplify notation. The objective is then equivalent to 
    $\sum_{i = 1}^{d_\text{row}} || \mathbf{W_{i,:}} \mathbf{X} - \mathbf{\widehat{W}_{i, :}} \mathbf{X}||_2^2.$
    
This way of writing the error makes it clear that removing a single weight $[\mathbf{W}]_{ij}$ only affects the error of the corresponding output row $\mathbf{Y_{i, :}} = \mathbf{W_{i, :}}\mathbf{X}$. Hence, there is no Hessian interaction between different rows and so it suffices to work only with the individual $d_\text{col} \times d_\text{col}$ Hessians corresponding to each of the $d_\text{row}$ rows. Further, as the dense layer output $\mathbf{Y} = \mathbf{W} \mathbf{X}$ is fixed, the objective for each row has standard least squares form and its Hessian is given by $\mathbf{H} = \mathbf{2\mathbf{X}\mathbf{X}^\top}$.

Although this observation already reduces computational complexity, two key challenges remain: (a) applying OBS to each row still costs $O(d_\text{col} \cdot d_\text{col}^3)$ time, which is too slow for large layers, and (b) we need fast access to the Hessian inverses of all $d_\text{row}$ rows, since we want to prune the minimum score weight of the whole matrix rather than just per row in each step. In particular, (b) requires $O(d_\text{row} \cdot d_\text{col}^2)$ GPU memory, which is likely to be infeasible.

\paragraph{Step 1: Handling a Single Row.} We first describe how to efficiently prune weights from a single row with $d_\text{col}$ parameters. For simplicity, we denote such a row by $\mathbf{w}$ with corresponding Hessian $\mathbf{H}$. 
The full algorithm for this procedure is given in Algorithm~\ref{alg:trueobs-row}; in the following, we provide a detailed description. 
The key idea is to avoid having to do the full $\Theta(N \cdot d_\text{col}^2)$ calculation and $\Theta(d_\text{col}^3)$ inversion of $\mathbf{H}$ in each step.
The former is easy, as the weights themselves do not enter the calculation of $\mathbf{H} = 2\mathbf{X}\mathbf{X}^\top$, and the Hessian for the weights with pruning mask $M$ denoted by $\mathbf{H}_M$ is thus simply comprised of the corresponding rows and columns in the fully dense version $\mathbf{H}$. Hence, we only have to compute $\mathbf{H}$ (which is actually the same for all rows) once, from which we can then extract the rows and columns corresponding to $M$ as needed. 

Critically, this trick is \textit{not} applicable to the inverse, as $(\mathbf{H}_M)^{-1} \neq (\mathbf{H}^{-1})_M$. However, using the fact that the removal of one parameter $p$ simply drops the corresponding row and column from $\mathbf{H}$, we can actually update the inverse to remove parameter $p$ directly using a single step of Gaussian elimination, with cost $\Theta(d_{col}^2)$.
The following result, whose proof is in the Appendix, formalizes this.

\begin{lemma}[Row \& Column Removal]
\label{lem:row-element}
Given an invertible $d_\text{col} \times d_\text{col}$ matrix $\mathbf{H}$ and its inverse $\mathbf{H}^{-1}$, we want to efficiently compute the inverse of $\mathbf{H}$ with row and column $p$ removed, which we denote by $\mathbf{H}_{-p}$. This can be accomplished through the following  formula:
\begin{equation}
    \mathbf{H}_{-p}^{-1} = \Big(\mathbf{H}^{-1} - \frac{1}{[\mathbf{H}^{-1}]_{pp}} \mathbf{H}^{-1}_{:, p} \mathbf{H}^{-1}_{p, :} \Big)_{-p},
\end{equation}
which corresponds to performing Gaussian elimination of row and column $p$ in $\mathbf{H}^{-1}$ followed by dropping them completely. This has $\Theta(d_\text{col}^2)$ time complexity.
\end{lemma}

The resulting pseudocode is shown in Algorithm \ref{alg:trueobs-row}, where we avoid constantly resizing $\mathbf{H}^{-1}$ (and correspondingly changing indices) by utilizing the fact that row and column $p$ have no effect on any future calculations after they have been eliminated by Lemma \ref{lem:row-element} as they are 0 (and the non-zero diagonal element is never accessed again).
One can check that this algorithm applies OBS to a single row of $\mathbf{W}$  with a per-step cost of $\Theta(d_\text{col}^2)$, and thus $\Theta(k \cdot d_\text{col}^2)$ overall time for pruning $k$ weights.

\begin{minipage}{0.475\textwidth}
    \begin{algorithm}[H]
        \centering
        \caption{Prune $k \leq d_\text{col}$ weights from row $\mathbf{w}$ with inverse Hessian $\mathbf{H}^{-1} = (2 \mathbf{X} \mathbf{X}^\top)^{-1}$ according to OBS in $O(k \cdot d_\text{col}^2$) time.}
        \label{alg:trueobs-row}
        \begin{algorithmic}
            \STATE $M = \{1, \dots, d_\text{col}\}$
            \FOR {$i = 1, \dots, k$}
                \STATE $p \gets \text{argmin}_{p \in M} \frac{1}{[\mathbf{H}^{-1}]_{pp}} \cdot w_p^2$
                \STATE $\mathbf{w} \gets \mathbf{w} - \mathbf{H}^{-1}_{:, p} \frac{1}{[\mathbf{H}^{-1}]_{pp}} \cdot w_p$
                \STATE $\mathbf{H}^{-1} \gets \mathbf{H}^{-1} - \frac{1}{[\mathbf{H}^{-1}]_{pp}} \mathbf{H}^{-1}_{:, p} \mathbf{H}^{-1}_{p, :}$
                \STATE $M \gets M - \{p\}$
            \ENDFOR
        \end{algorithmic}
    \end{algorithm}
    \vspace{5pt}
\end{minipage}
\hfill
\begin{minipage}{0.475\textwidth}
    \centering
    \includegraphics[width=\linewidth]{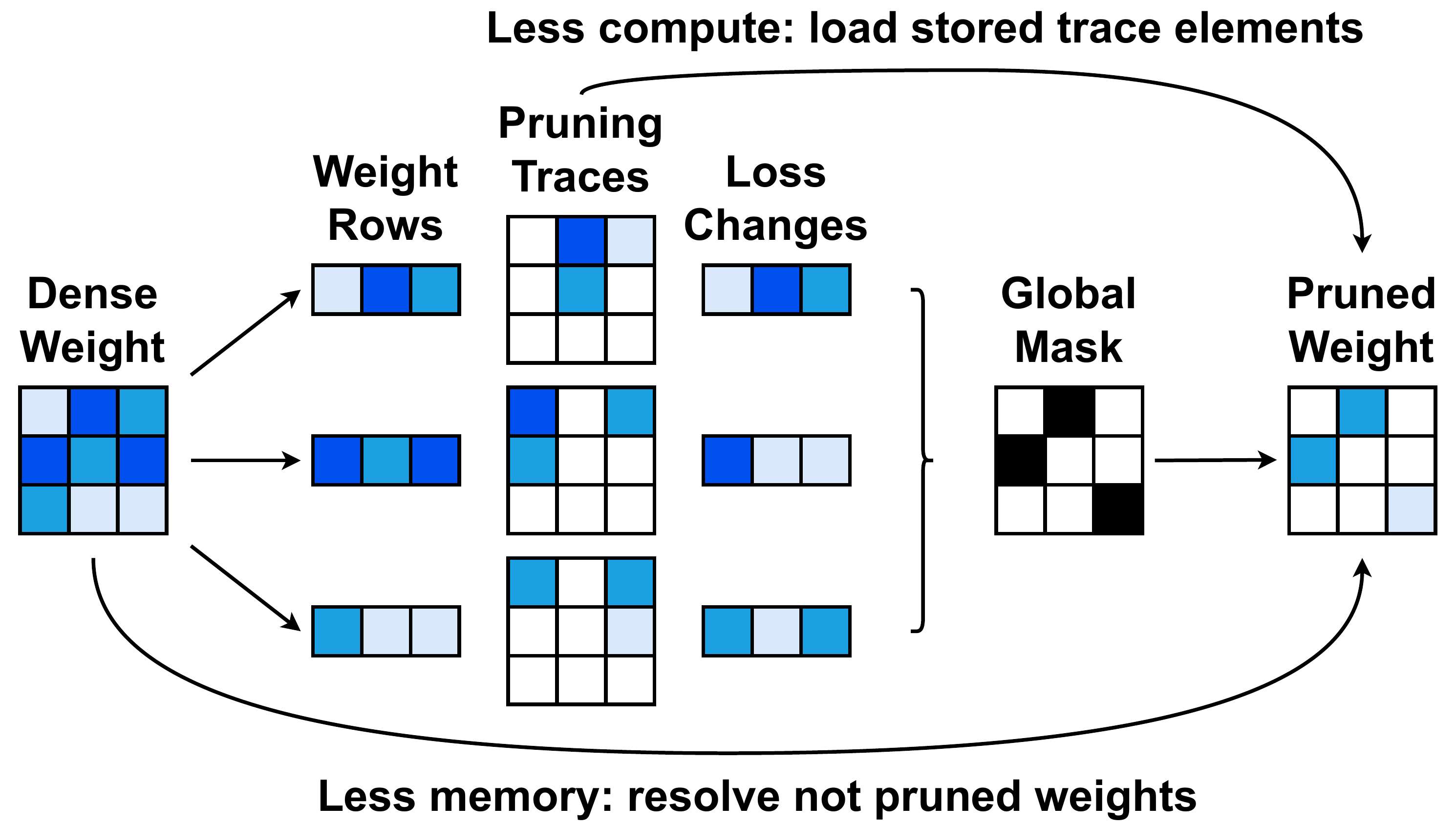}
    \captionof{figure}{Efficient global OBS using the row-wise results.}
    \label{fig:global-trueobs}
\end{minipage}

\paragraph{Step 2: Jointly Considering All Rows.} Applying the OBS framework to the full weight matrix $\mathbf{W}$ rather than just to each row independently requires fast access to all $d_\text{row}$ row-wise inverse Hessians, in order to select the weight with the  smallest overall pruning score in each step. 
However, storing $d_\text{row}$ matrices of size $d_\text{col} \times d_\text{col}$ each in GPU memory can be too expensive; while it would be possible to offload some Hessians to main memory, this could  result in a large number of expensive memory transfers.
However, since there is no Hessian interaction between rows, the final compressed weights of each row only depend on the total number of parameters that were pruned in it. Similarly, the change in loss incurred by pruning some weight only depends on the previously pruned weights in the same row, which also means that the order in which weights are pruned in each row is fixed.

The consequence of these insights is that we can process each row independently, pruning all weights in order while always recording the corresponding change in loss $\delta \mathcal{L}_p = w_p^2 / [\mathbf{H}^{-1}]_{pp}$. At the end, we know $\delta \mathcal{L}_p$ for all $d$ weights and can then simply determine the global mask that would be chosen by OBS on the full matrix by selecting the weights with the lowest values in order, requiring only $\Theta(d)$ extra memory. We note that once the per-row masks $M_i$ are known, 
we can directly solve for the optimal update of the remaining weights via the corresponding group OBS formula \cite{kurtic2022optimal} $\boldsymbol{\delta_{M_i}} = \mathbf{H}^{-1}_{:, {M_i}}((\mathbf{H}^{-1})_{M_i})^{-1}\mathbf{w}_{M_i}$. This will be considerably faster in practice than simply rerunning the iterative pruning process in Algorithm \ref{alg:trueobs-row}. Alternatively, if enough CPU memory is available, one can keep the full \textit{pruning trace} of each row, that is, the full weight vector after every individual pruning step, in CPU memory and ultimately simply reload the entries corresponding to the global mask. This requires $O(d_\text{row} \cdot d_\text{col}^2)$ extra CPU memory but avoids a second computation pass to reconstruct the not pruned weights and will therefore be faster. Figure \ref{fig:global-trueobs} visualizes both options just discussed.

\paragraph{Implementation Details.} In practice, the matrix $\mathbf{H}$ might not always be invertible for reasons such as using too few data samples or dead / linearly dependent inputs. The former can usually be addressed by extending the calibration dataset with augmentations (additional augmented samples only need to be accumulated into the Hessian once and are thus very cheap to include) and the latter can be prevented by adding a small diagonal dampening term to the Hessian before inverting it. 
Second, a direct GPU implementation of Algorithm \ref{alg:trueobs-row} will perform a large number of small CUDA calls, which can be expensive.
This overhead can be removed by using batch operations to process multiple matrix rows simultaneously---for more details please see our sample implementation. Finally, when applied to an already sparse weight matrix, the complexity of our algorithm can scale cubicly with the row-density by working with a dense version of the weights / Hessians consisting only of the non-zero elements and mapping the pruning result back at the end.

\paragraph{N:M Sparsity.}
Our method can be easily extended to various forms of \textit{semi-structured} sparsity. This includes, for example, the N:M sparsity pattern \cite{zhou2021learning}, which enforces exactly $N$ non-zero values in each block of $M$ consecutive weights, and is becoming popular due to support on newer NVIDIA hardware~\cite{NVIDIASparse}. Adapting our algorithm to this pattern requires only one simple change: instead of selecting the weight with the smallest change in loss, we select the weight with the smallest change in loss that is in a block with $< N$ pruned weights. We note that all rows have exactly the same sparsity $1 - N/M$ in the N:M pattern and so we can terminate per-row pruning as soon as this target sparsity value is reached. For the same reason, there is no need for the global mask selection step described earlier. Thus, our method will be even more efficient in this case.

\paragraph{Block-Sparsity.}
Another practically relevant pruning pattern, particularly in the context of CPU acceleration~\cite{elsen2020fast, pmlr-v119-kurtz20a}, is \emph{block-pruning}, where zeros appear only in consecutive blocks of size $c$, which is typically a small number like 4 or 8. We follow recent work \cite{kurtic2022optimal} that extends the OBS framework to pruning small groups of connected weights in order to account for the correlation between them, using the following formulas for the target block and weight update, respectively:
\begin{equation}
    \mathbf{w}_P = \text{argmin}_{\mathbf{w}_P} \, \mathbf{w}_P^\top ((\mathbf{H}^{-1})_P)^{-1} \mathbf{w}_P, \quad \boldsymbol{\delta_P} = - \mathbf{H}^{-1}_{:, P} ((\mathbf{H}^{-1})_P)^{-1} \mathbf{w}_P,
\end{equation}
where $P$ denotes the set of indices corresponding to one block. Algorithm \ref{alg:trueobs-row} can easily be adapted to operate on blocks using the above equations and applying the update of $\mathbf{H}^{-1}$ via Lemma~\ref{lem:row-element} successively for all $p \in P$. Although there are now only $d_{col} / c$ steps per row, each update of $\mathbf{H^{-1}}$ also takes $O(c \cdot d_\text{col}^2)$ time and so the overall asymptotic runtime stays the same. Additional practical overhead only comes from the extra $O(c^2 \cdot d_\text{col}^2)$ terms that are the result of computing and multiplying with the $c \times c$ matrices $((\mathbf{H}^{-1})_P)^{-1}$.

\section{The Optimal Brain Quantizer (OBQ)}

Although the classical OBS framework~\cite{lecun1990optimal, hassibi1993optimal} has inspired a long line of work on pruning methods for DNNs~\cite{singh2020woodfisher, frantar2021m, liu2021group}, 
so far it has not been used for quantization.
We now show that our results from the previous section can in fact be extended to quantization in an effective and accurate way, via a method which we call the Optimal Brain Quantizer (OBQ), in the spirit of~\cite{lecun1990optimal, hassibi1993optimal}. 

\paragraph{The Quantization Order and Update Derivations.} 
Under the standard assumption that the gradient at the current point $\mathbf{w}$ is negligible, the OBS formulas for the optimal weight to be pruned $w_p$ and the corresponding update $\boldsymbol{\delta_p}$ can be derived by writing the locally quadratic problem under the constraint that element $p$ of $\boldsymbol{\delta_p}$ is equal to $-w_p$, which means that $w_p$ is zero after applying the update to $\mathbf{w}$. This problem has the following Lagrangian:
\begin{equation}
    \label{eq:obs-lagrangian}
    L(\boldsymbol{\delta_p}, \lambda) = \boldsymbol{\delta}^\top_{\mathbf{p}} \mathbf{H} \boldsymbol{\delta_p} + \lambda (\mathbf{e}^\top_{\mathbf{p}} \boldsymbol{\delta_p} - (-w_p)),
\end{equation}
where $\mathbf{H}$ denotes the Hessian at $\mathbf{w}$ and $\mathbf{e_p}$ is the $p$th canonical basis vector. The optimal solution is then derived by first finding the optimal solution to $\boldsymbol{\delta_p}$ via setting the derivative $\partial L / \partial \boldsymbol{\delta_p}$ to zero and then substituting this solution back into $L$ and solving for $\lambda$; please see e.g. \cite{hassibi1993optimal, singh2020woodfisher} for examples.

Assume a setting in which we are looking to quantize the weights in a layer on a fixed grid of width $\Delta$ while minimizing the loss. 
To map OBS to a \emph{quantized} projection, we can set the target of the Lagrangian constraint in (\ref{eq:obs-lagrangian}) to $(\text{quant}(w_p) - w_p)$, 
where $\text{quant}(w_p)$ is the weight rounding given by quantization; then $w_p = \text{quant}(w_p)$ after the update.

Assuming we wish to quantize weights iteratively, one-at-a-time, we can derive formulas for the ``optimal'' weight to quantize at a step, in terms of minimizing the loss increase, and for the corresponding optimal update to the unquantized weights, in similar fashion as discussed above:
\begin{equation}
    \label{eq:obs-quant}
    w_p = \text{argmin}_{w_p} \, \frac{(\text{quant}(w_p) - w_p)^2}{[\mathbf{H}^{-1}]_{pp}}, \quad \boldsymbol{\delta_p} = - \frac{w_p - \text{quant}(w_p)}{[\mathbf{H}^{-1}]_{pp}} \cdot \mathbf{H}^{-1}_{:, p}.
\end{equation}
In fact, since $-w_p$ is a constant during all derivations, we can just substitute it with $(\text{quant}(w_p) - w_p)$ in the final result. We note that the resulting formulas are a generalization of standard OBS for pruning, if $\text{quant}(\cdot)$ always ``quantizes'' a weight to 0, then we recover the original form.

\paragraph{Quantizing Full Layers.} 
At first glance, OBQ might appear curious since one usually quantizes \textit{all} weights in a layer, leaving no more weights to update. 
At the same time, the weight selection metric influences only the quantization order, but not the quantization value. 
However, this view changes when considering OBQ in the context of our efficient one-weight-at-a-time pruning algorithm described in the previous section. Specifically, using OBQ, we can greedily quantize the currently ``easiest'' weight by the above metric, and then adjust all the remaining unquantized weights to compensate for this loss of precision, thus changing their value. We then choose the next weight to quantize, and so on. 
This can result in quantization assignments that are different from the ones that would have been chosen by rounding initially, and in better overall quantization results. 
Concretely, to realize this, we can plug (\ref{eq:obs-quant}) into Algorithm~\ref{alg:trueobs-row} in order to iteratively quantize weights for a given layer, leading to the similar Algorithm in the Appendix, thus essentially unifying pruning and quantization.

\paragraph{Quantization Outliers.}
One practical issue with this greedy scheme can occur especially when applied to quantization grids that permit some outliers in order to achieve a lower error on the majority of weights, which are currently standard \cite{choukroun2019low, nahshan2021loss}. Since these outliers can have high quantization error, they will usually be quantized last, when there are only few other unquantized weights available that may be adjusted to compensate for the large error incurred by quantizing the outliers. This effect can become worse when some weights are pushed even further outside the grid by intermediate updates. We prevent this with a simple but effective heuristic: we quantize outliers, e.g. weights with a quantization error $> \Delta / 2$ where $\Delta$ is the distance between quantized values, as soon as they appear (which typically happens only a few times per layer). With this heuristic, OBQ yields a highly effective layer-wise quantization scheme, as our experiments in the next section demonstrate. Finally, we note that the OBQ version of the techniques discussed in Section \ref{sec:trueobs} has all the same runtime and memory characteristics (barring the global step in Figure \ref{fig:global-trueobs}, which is unnecessary for quantization).

\section{Experiments}
\label{sec:experiments}

\paragraph{Objectives, Models \& Datasets.} To demonstrate the effectiveness and flexibility of our method, we consider several different standard \emph{post-training compression} scenarios~\cite{nagel2020up, hubara2021accurate, hubara2021accelerated}. We begin with settings where only a single type of compression is applied: concretely, we consider unstructured pruning for given FLOP targets, global 2:4 and 4:8 pruning, as well as uniform weight quantization. Additionally, we also study two practical tasks that feature joint pruning and quantization: a GPU scenario where quantization and N:M pruning are combined, as well as a CPU scenario combining quantization and block pruning. We work with variants of the following models and tasks: ResNet~\cite{he2016deep} for image classification on Imagenet \cite{imagenet}, YOLOv5 \cite{yolov5} for object detection on COCO \cite{lin2014microsoft} and BERT \cite{devlin2018bert} for question answering on SQuAD \cite{rajpurkar2016squad}. Our smaller BERT models denoted by BERT3 and BERT6 correspond to the smaller 3 and 6 layer variants of BERT-base, respectively, trained by \cite{kurtic2022optimal}. The Appendix contains additional experiments as well as runtime information of our algorithms.

\paragraph{Experimental Setup.}
All of our calibration datasets consist of 1024 random training samples. For ImageNet, where we use roughly $0.1\%$ of the training data, we additionally apply standard flipping and cropping augmentations to artificially increase the size of this dataset by $10\times$; other tasks do not use any augmentations. While the effect of  augmentations is typically minor, they are very cheap to include for our method. For ResNet models, batchnorm statistics are reset using 100 batches of 128 samples from the calibration set with standard augmentations. For other models, we apply mean and variance correction \cite{nagel2019data, banner2019post} after all normalization layers (so that the correction parameters can be easily merged and incur no extra cost) on a single batch of samples of size 128 (for YOLO) and 512 (for BERT). We found this to be more effective than batchnorm tuning for YOLO, and the BERT models have no batchnorm layers.

When compressing to a given FLOP or timing constraint, we need to solve the problem of identifying per-layer compression targets, which match the constraint, while maximizing accuracy. 
To identify these non-uniform targets, we follow the approach of \cite{frantar2021m}: we first collect a ``model database'' containing for each compression level (e.g. bit-width or sparsity setting) the corresponding (independently) compressed version of each layer. For building a joint sparse and quantized database we simply sparsify layers first and then apply quantization to the remaining weights.
Next, similarly to~\cite{hubara2021accurate}, we compute the layer-wise calibration losses (without augmentations) for all compression levels, corresponding to the models with exactly one layer compressed to a certain level. 
Then, given layer-wise FLOP or timing information, we set up a constrained layer-wise compression problem of the form described in AdaQuant \cite{hubara2021accurate} and solve it with the dynamic programming algorithm of SPDY \cite{frantar2021m}. This returns an optimal per-layer assignment of compression levels, for which we can then easily produce the corresponding model, via a two-step process: 
we first stitch together layers at the corresponding compression levels from the database, and then perform the discussed statistics correction to recover some extra accuracy \cite{hubara2021accurate}.

\paragraph{Unstructured Sparsity.} We begin our experiments with \textit{unstructured} sparsity, comparing against global magnitude pruning (GMP) \cite{zhu2017prune}, the approximate layer-wise OBS method L-OBS \cite{2017-dong}, and the post-training pruning state-of-the-art method AdaPrune \cite{hubara2021accelerated}. As a sanity check, we examine in Figure \ref{fig:squared-error} whether our method provides better results in terms of layer-wise squared error, pruning the first layer of a ResNet18 (RN18) model to several sparsities. In this metric, ExactOBS performs best by a wide margin ahead of AdaPrune, which significantly outperforms the other two methods.

Next, in Table~\ref{tab:unstr}, we turn our attention to the practical problem of pruning various models to achieve a given FLOP reduction of $2\times$--$4\times$, applying the per-layer target sparsity optimization technique described above. Our ExactOBS generally performs best (except for YOLOv5l $2\times$ where all methods perform similarly in terms of mAP@0.5) and at $4\times$ FLOP reduction even with a $> 1\%$ gap to the next best method. Interestingly, on the hard-to-prune BERT model, ExactOBS appears to be the only method which still produces reasonable results at higher reduction targets. For BERT $3\times$ and $4\times$, where the performance drop of all methods is $> 2\%$, we additionally assess the compatibility of our results with the more powerful (but also more expensive) post processing method \textit{global AdaPrune}~\cite{frantar2021m}. While this global optimization technique is able to recover  lost accuracy, the ExactOBS models still maintain a $> 0.5\%$ and $> 2\%$ F1 advantage, respectively (see Table~\ref{tab:gap}).

\begin{table}[h!]
    \vspace{-5pt}
    \begin{minipage}[c]{.28\textwidth}
        \centering
        \includegraphics[width=\textwidth]{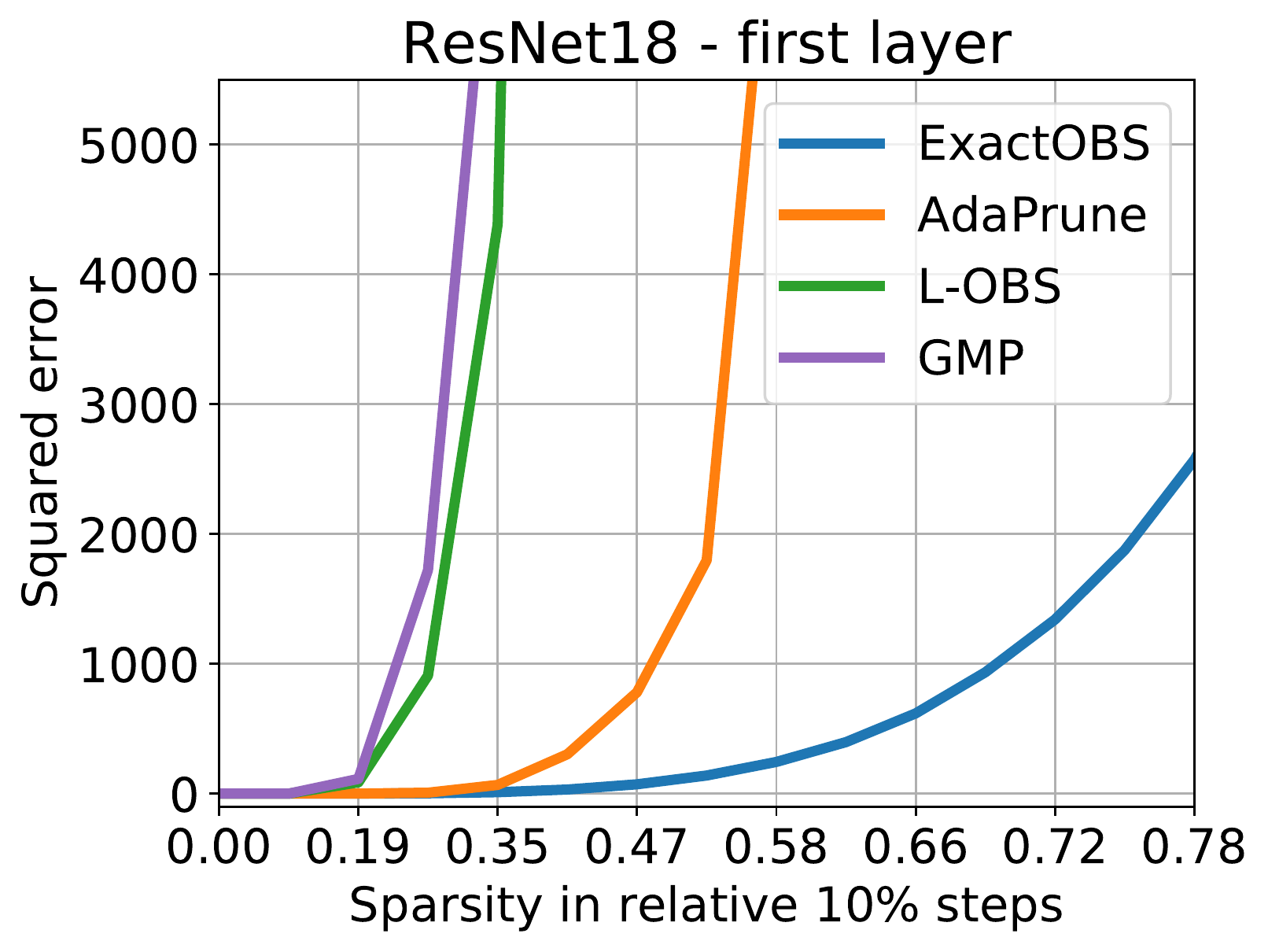}
        \renewcommand{\tablename}{Figure}
        \vspace{-10pt}
        \setcounter{table}{1}
        \caption{\small{RN18 squared error.}}
        \renewcommand{\tablename}{Table}
        \label{fig:squared-error}
    \end{minipage}
    \begin{minipage}[c]{.68\textwidth}
        \centering
        \scalebox{.75}{
            \begin{tabular}{|l|ccc|ccc|ccc|}
                \toprule
                \multirow{2}{*}{Method} & \multicolumn{3}{c|}{ResNet50 -- 76.13} & \multicolumn{3}{c|}{YOLOv5l -- 66.97} & \multicolumn{3}{c|}{BERT -- 88.53} \\
                & $2\times$ & $3\times$ & $4\times$ & $2\times$ & $3\times$ & $4\times$ & $2\times$ & $3\times$ & $4\times$ \\
                \midrule
                GMP & 74.86 & 71.44 & 64.84 & 65.83 & 62.30 & 55.09 & 65.64 & 12.52 & 09.23 \\
                L-OBS & 75.48 & 73.73 & 71.24 & \textbf{66.21} & 64.47 & 61.15 & 77.67 & 3.62 & 6.63 \\
                AdaPrune & 75.53 & 74.47 & 72.39 & 66.00 & 64.88 & 62.71 & 87.12 & 70.32 & 18.75 \\
                \midrule
                ExactOBS & \textbf{75.64} & \textbf{75.01} & \textbf{74.05} & 66.14 & \textbf{65.35} & \textbf{64.05} & \textbf{87.81} & \textbf{85.87} & \textbf{82.10} \\
                \bottomrule
            \end{tabular}
        }
        \vspace{5pt}
        \setcounter{table}{0}
        \caption{\small{Unstructured pruning for different FLOP reduction targets.}}
        \label{tab:unstr}
    \end{minipage}
    \vspace{-10pt}
\end{table}

\paragraph{N:M Sparsity.} Next, we study the performance of our method for \textit{semi-structured} sparsity via the N:M pattern. Specifically, we compare against the 4:8 results of AdaPrune with batchnorm tuning~\cite{hubara2021accelerated} on ResNet models (see Table~\ref{tab:nm-resnets}) in addition to a 2:4 comparison on BERT models (see Table~\ref{tab:nm-bert}). We highlight that ExactOBS matches or even slightly exceeds the 4:8 results of AdaPrune with the considerably more stringent 2:4 pattern, which is already well supported on NVIDIA hardware. Furthermore, in a 2:4 comparison on BERT models, ExactOBS achieves $1$--$2\%$ higher F1 scores.

\begin{table}[h!]
    \vspace{-2.5pt}
    \begin{minipage}[c]{.53\textwidth}
        \centering
        \scalebox{.75}{
            \begin{tabular}{|l|c|c|cc|}
                \toprule
                \multirow{2}{*}{Model} & \multirow{2}{*}{Dense} & AdaPrune & \multicolumn{2}{c|}{ExactOBS} \\
                & & 4:8 & 2:4 & 4:8 \\
                \midrule
                ResNet18 & 69.76 & 68.63 & 68.81 & \textbf{69.18} \\
                ResNet34 & 73.31 & 72.36 & 72.66 & \textbf{72.95} \\
                ResNet50 & 76.13 & 74.75 & 74.71 & \textbf{75.20} \\
                \bottomrule
            \end{tabular}
        }
        \vspace{5pt}
        \caption{\small{Semi-structured N:M pruning (+ batchnorm tuning) of all layers except the first and the last.}}
        \label{tab:nm-resnets}
    \end{minipage}
    \hfill
    \begin{minipage}[c]{.43\textwidth}
        \centering
        \scalebox{.75}{
            \begin{tabular}{|l|c|c|c|}
                \toprule
                Model & Dense & AdaPrune & ExactOBS \\
                \midrule
                BERT3 & 84.66 & 82.75 & \textbf{83.54} \\
                BERT6 & 88.33 & 85.02 & \textbf{86.97} \\
                BERT & 88.53 & 85.24 & \textbf{86.77} \\ 
                \bottomrule
            \end{tabular}
        }
        \vspace{5pt}
        \caption{\small{Semi-structured 2:4 pruning of all layers except the embeddings.}}
        \label{tab:nm-bert}
    \end{minipage}
\end{table}

\paragraph{Quantization.} Additionally, we compare OBQ's \textit{independent} performance (after batchnorm tuning) with the state-of-the-art \textit{sequential} post-training methods AdaQuant \cite{hubara2021accurate}, AdaRound \cite{nagel2020up} and BRECQ~\cite{li2021brecq}. We perform standard asymmetric per-channel quantization of all weights, using the authors' implementations. We rerun all methods on Torchvision \cite{marcel2010torchvision} ResNets to ensure a uniform baseline. The quantization grids for OBQ as well as AdaRound are determined with the same LAPQ \cite{nahshan2021loss} procedure that is used by BRECQ. Surprisingly, we find that, despite optimizing layers independently, OBQ achieves very similar (sometimes even slightly better) accuracies as existing non-independent methods for 4 and 3 bits. This suggests that it should be well-suited for mixed precision applications where one needs to quickly generate many non-uniform models optimized for different constraints. (However, we note that ExactOBS can also be applied sequentially; see Appendix.)

\begin{table}[ht!]
    \begin{minipage}[c]{.65\textwidth}
        \centering
        \scalebox{.75}{
            \begin{tabular}{|l|cc|ccc|ccc|}
                \toprule
                \multirow{2}{*}{Method} & \multirow{2}{*}{Lw.} & \multirow{2}{*}{Ind.} & \multicolumn{3}{c|}{ResNet18 -- 69.76} & \multicolumn{3}{c|}{ResNet50 -- 76.13} \\
                & & & 4bit & 3bit & 2bit & 4bit & 3bit & 2bit \\
                \midrule
                AdaRound & yes & no & 69.34 & 68.37 & 63.37 & 75.84 & 75.14 & 71.58 \\
                AdaQuant & yes & no & 68.12 & 59.21 & 00.10 & 74.68 & 64.98 & 00.10 \\
                BRECQ & no & no & 69.37 & 68.47 & 64.70 & 75.88 & 75.32 & 72.41 \\
                \midrule
                OBQ (ours) & yes & yes & 69.56 & 68.69 & 64.04 & 75.72 & 75.24 & 70.71 \\
                \bottomrule
            \end{tabular}
        }
        \vspace{5pt}
        \caption{\small{Comparison with state-of-the-art post-training methods for asymmetric per-channel weight quantization of all layers. We mark whether methods are Layer-wise (Lw.) or Independent (Ind.).}}
        \label{tab:quant-seq}
    \end{minipage}
    \hfill
    \begin{minipage}[c]{.32\textwidth}
        \centering
        \scalebox{.75}{
            \begin{tabular}{|l|cc|}
                \toprule
                \multirow{2}{*}{Methods} & \multicolumn{2}{c|}{BERT} \\
                & $3\times$ & $4\times$ \\
                \midrule
                gAP + AdaPrune & 86.99 & 84.10 \\
                gAP + ExactOBS & \textbf{87.57} & \textbf{86.42} \\
                \bottomrule
            \end{tabular}
        }
        \vspace{5pt}
        \caption{\small{Further improving results in Table \ref{tab:unstr} with $> 3\%$ performance drops through more expensive post-processing via global AdaPrune (gAP).}}
        \label{tab:gap}
    \end{minipage}
    \vspace{-15pt}
\end{table}
\begin{figure}[h!]
    \centering
    \begin{subfigure}{.4\textwidth}
      \centering
      \includegraphics[width=\linewidth]{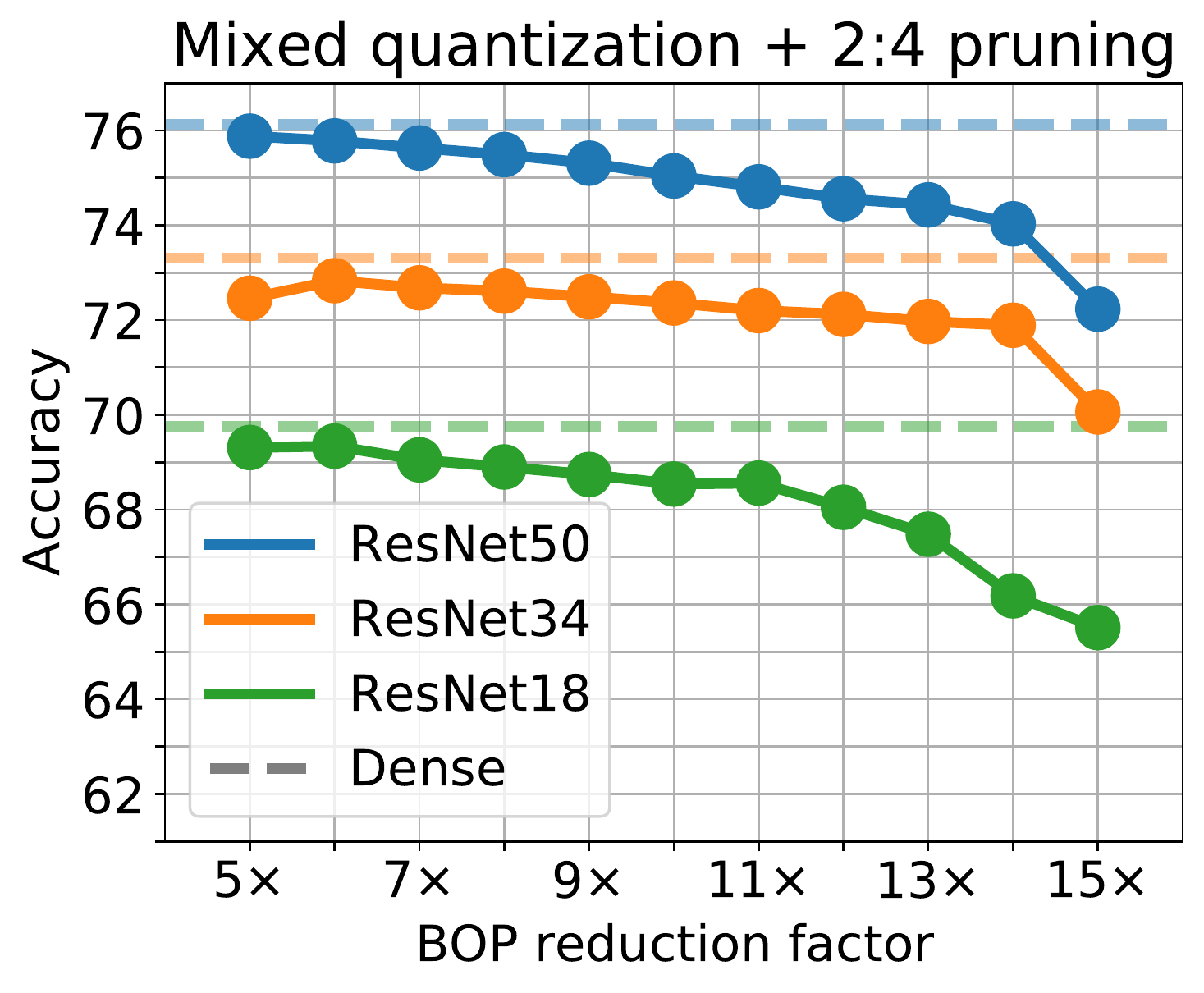}
      \vspace{-15pt}
      \caption{ResNet variants.}
      \label{fig:rn-mixed}
    \end{subfigure}~~~~~~
    \begin{subfigure}{.4\textwidth}
      \centering
      \includegraphics[width=\linewidth]{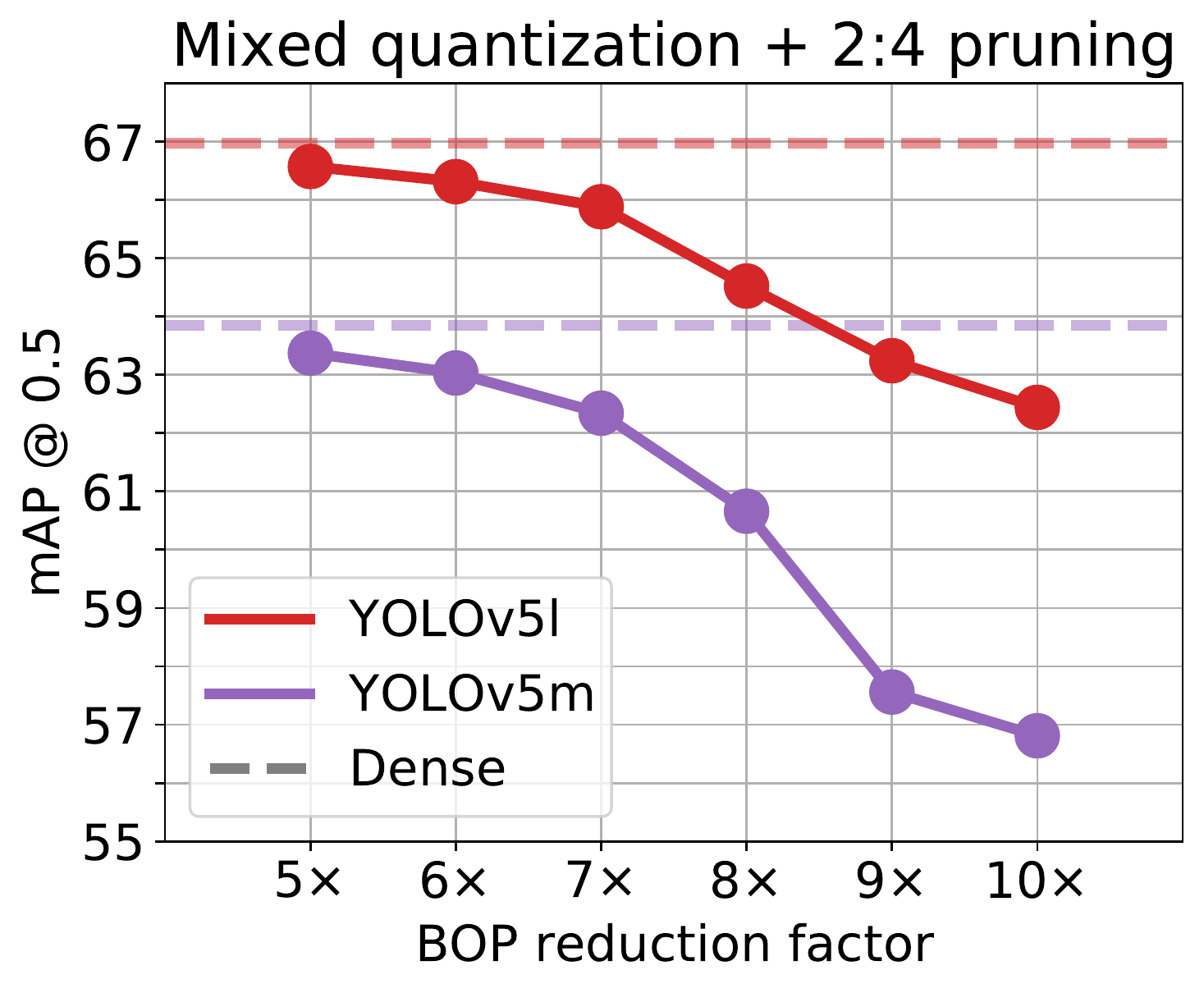}
      \vspace{-15pt}
      \caption{YOLO variants.}
      \label{fig:yolo-mixed}
    \end{subfigure}\\
    \begin{subfigure}{.4\textwidth}
      \centering
      \includegraphics[width=\linewidth]{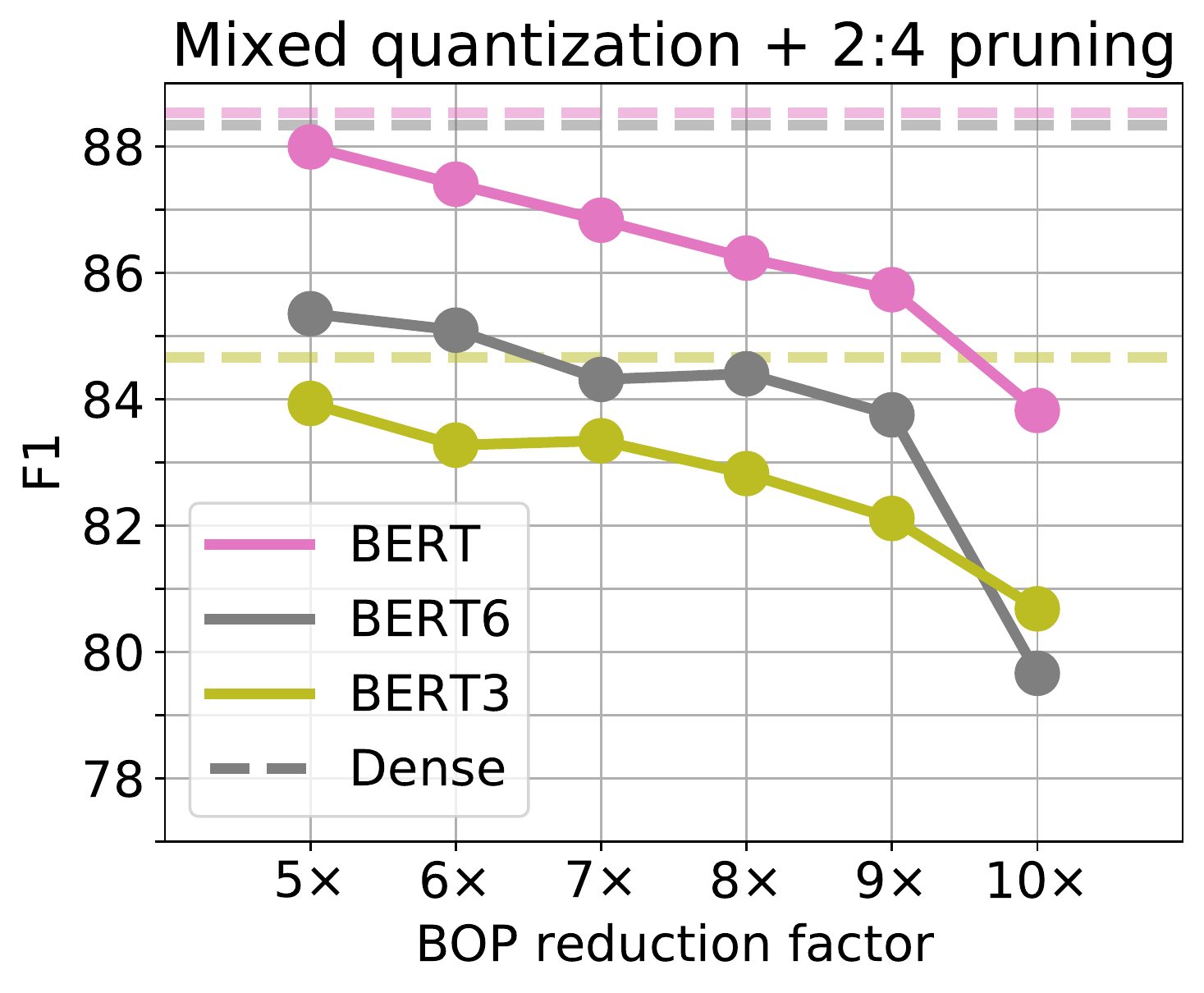}
      \vspace{-15pt}
      \caption{BERT variants.}
      \label{fig:bert-mixed}
    \end{subfigure}~~~~~~
    \begin{subfigure}{.4\textwidth}
      \centering
      \includegraphics[width=\linewidth]{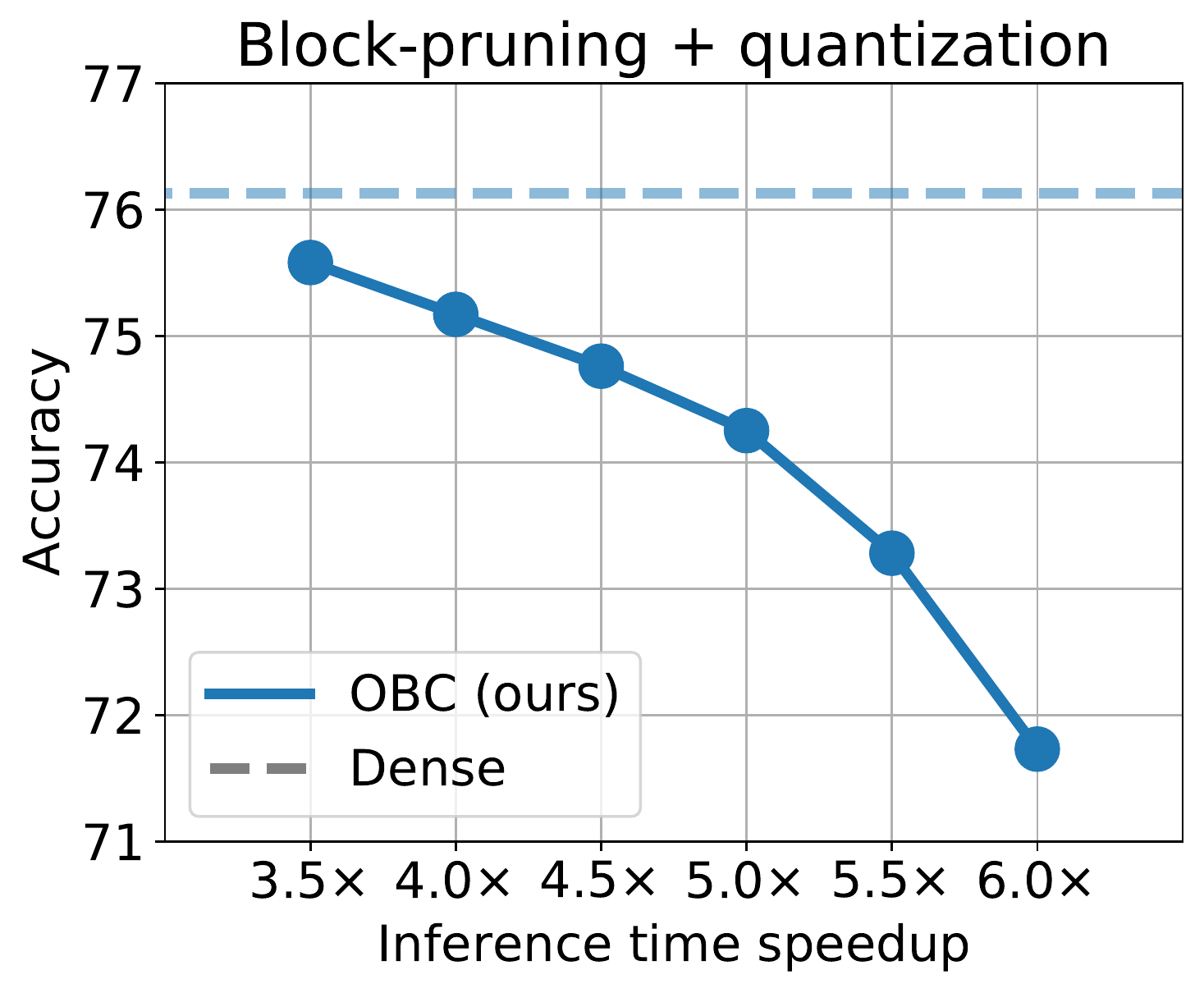}
      \vspace{-15pt}
      \caption{ResNet50 -- CPU.}
      \label{fig:rn-cpu}
    \end{subfigure}
    \setcounter{figure}{3}
    \caption{(a) to (c) Mixed quantization and 2:4 pruning for various BOP reduction targets. (d) Joint block-pruning and quantization for CPU inference time speedups.}
    \label{fig:mixed}
\end{figure}

\paragraph{BOP-Constrained Mixed GPU Compression.}
We now consider a practical setting where we are given a trained model together with some calibration data and want to compress this model for efficient inference on an NVIDIA GPU which supports 8-bit and 4-bit arithmetic, also in combination with 2:4 sparsity. Thus, there are 4 possible compression choices per layer: 8bit weights + 8bit activations (8w8a), 4w4a, 8w8a + 2:4 and 4w4a + 2:4. Unlike in the previous section, we do \textit{symmetric} per-channel quantization of the weights as it has better hardware support; activations are quantized asymmetrically per-tensor. We then generate mixed precision configurations for various BOP (number of bits times FLOPs) reduction targets and visualize the resulting compression-accuracy trade-off curves in Figure~\ref{fig:mixed}. In summary, at the cost of a $\approx 2.5\%$ relative performance drop, we can achieve a $12 - 14\times$ BOP reduction for ResNets and a $7 - 8\times$ reduction for the more challenging YOLO and BERT models (relative to the compute in compressible layers). To the best of our knowledge, we are the first to consider joint N:M pruning and quantization in a post-training setting. Recent work \cite{chmiel2022optimal} also studies joint 4w4a + 2:4 compression for ResNet18 but with 90 epochs of (sparse) Quantization-Aware Training (QAT) on the full dataset and report 67.33\% accuracy. Although not perfectly comparable (we keep the first layer dense and their dense baseline has $0.94\%$ higher accuracy and uses 4:8 sparse activations), we achieve similar 67.20\% accuracy for 4w4a + 2:4 \textit{post training}, which  emphasizes the effectiveness of our methods for joint sparsification and quantization.

\paragraph{Time-Constrained CPU Compression.}
Lastly, we explore a similar scenario, but targeting actual CPU inference speedup on a 12-core Intel Xeon Silver 4214 CPU using the DeepSparse inference engine \cite{deepsparse, pmlr-v119-kurtz20a}, which provides acceleration for joint 8-bit quantization and block-sparsity with blocksize 4. In this case, we work with real layer-wise timing data (for batchsize 64), as in  \cite{frantar2022spdy}. There are 30 available block-sparsity targets per-layer, in steps of pruning 10\% of the remaining weights, all of which are further quantized to 8 bits. The base acceleration of the dense 8 bit model is $\approx 2.7\times$ on top of which sparsity speedup acts roughly multiplicatively. Figure \ref{fig:rn-cpu} shows results for ResNet50 and several (real-time) speedup targets---we achieve $4\times$ and $5\times$ (actual) speedup with $1\%$ and $2\%$ accuracy loss, respectively. These are the first full post-training results in this setting (the authors of \cite{frantar2021m} only performed 4-block pruning post-training, followed by 5 epochs of QAT on the entire ImageNet dataset), and they show very encouraging accuracy-speedup trade-offs.

\section{Conclusions \& Future Work}

We have presented a new efficient and accurate approach for solving the layer-wise compression problem, and built on it to obtain state-of-the-art post-training compression solutions for both pruning and quantization. Our framework should be naturally extensible to \emph{structured} pruning, which in fact should allow for further optimizations, and should also be compatible with further compression via unstructured pruning and quantization. Our results suggest that post-training compression may be able to reach comparable accuracies to much more expensive retraining methods. 
We plan to investigate this in future work, in particular in the context of more resource-intensive models, such as very large-scale language models.

\section{Acknowledgements}

We gratefully acknowledge funding from the European Research Council (ERC) under the European Union’s Horizon 2020 programme (grant agreement No 805223 ScaleML), as well as computational support from AWS EC2. We thank Eldar Kurtic for providing us BERT code and pretrained models, and the Neural Magic Team, notably Michael Goin and Mark Kurtz, for support with their software.

\bibliographystyle{plain}

\begin{thebibliography}{10}

\bibitem{banner2019post}
Ron Banner, Yury Nahshan, and Daniel Soudry.
\newblock Post training 4-bit quantization of convolutional networks for
  rapid-deployment.
\newblock In {\em Conference on Neural Information Processing Systems
  (NeurIPS)}, 2019.

\bibitem{blumensath2008iterative}
Thomas Blumensath and Mike~E Davies.
\newblock Iterative thresholding for sparse approximations.
\newblock {\em Journal of Fourier Analysis and Applications}, 14(5-6):629--654,
  2008.

\bibitem{chmiel2022optimal}
Brian Chmiel, Itay Hubara, Ron Banner, and Daniel Soudry.
\newblock Optimal fine-grained {N:M} sparsity for activations and neural
  gradients.
\newblock {\em arXiv preprint arXiv:2203.10991}, 2022.

\bibitem{choukroun2019low}
Yoni Choukroun, Eli Kravchik, Fan Yang, and Pavel Kisilev.
\newblock Low-bit quantization of neural networks for efficient inference.
\newblock In {\em International Conference on Computer Vision Workshop
  (ICCVW)}, 2019.

\bibitem{devlin2018bert}
Jacob Devlin, Ming-Wei Chang, Kenton Lee, and Kristina Toutanova.
\newblock {BERT}: Pre-training of deep bidirectional transformers for language
  understanding.
\newblock In {\em North American Chapter of the Association for Computational
  Linguistics (NAACL)}, 2019.

\bibitem{2017-dong}
Xin Dong, Shangyu Chen, and Sinno~Jialin Pan.
\newblock Learning to prune deep neural networks via layer-wise optimal brain
  surgeon.
\newblock In {\em Conference on Neural Information Processing Systems
  (NeurIPS)}, 2017.

\bibitem{elsen2020fast}
Erich Elsen, Marat Dukhan, Trevor Gale, and Karen Simonyan.
\newblock Fast sparse convnets.
\newblock In {\em Conference on Computer Vision and Pattern Recognition
  (CVPR)}, 2020.

\bibitem{evci2018mean}
Utku Evci, Nicolas Le~Roux, Pablo Castro, and Leon Bottou.
\newblock Mean replacement pruning.
\newblock 2018.

\bibitem{frantar2022spdy}
Elias Frantar and Dan Alistarh.
\newblock {SPDY:} {A}ccurate pruning with speedup guarantees.
\newblock In {\em International Conference on Machine Learning (ICML)}, 2022.

\bibitem{frantar2021m}
Elias Frantar, Eldar Kurtic, and Dan Alistarh.
\newblock {M-FAC}: Efficient matrix-free approximations of second-order
  information.
\newblock In {\em Conference on Neural Information Processing Systems
  (NeurIPS)}, 2021.

\bibitem{gholami2021survey}
Amir Gholami, Sehoon Kim, Zhen Dong, Zhewei Yao, Michael~W Mahoney, and Kurt
  Keutzer.
\newblock A survey of quantization methods for efficient neural network
  inference.
\newblock {\em arXiv preprint arXiv:2103.13630}, 2021.

\bibitem{grosse2016kroneckerfactored}
Roger Grosse and James Martens.
\newblock A {K}ronecker-factored approximate {F}isher matrix for convolution
  layers.
\newblock In {\em International Conference on Machine Learning (ICML)}, 2016.

\bibitem{hassibi1993optimal}
Babak Hassibi, David~G Stork, and Gregory~J Wolff.
\newblock Optimal brain surgeon and general network pruning.
\newblock In {\em IEEE International Conference on Neural Networks}, 1993.

\bibitem{he2016deep}
Kaiming He, Xiangyu Zhang, Shaoqing Ren, and Jian Sun.
\newblock Deep residual learning for image recognition.
\newblock In {\em Conference on Computer Vision and Pattern Recognition
  (CVPR)}, 2016.

\bibitem{he2018amc}
Yihui He, Ji~Lin, Zhijian Liu, Hanrui Wang, Li-Jia Li, and Song Han.
\newblock {AMC}: {AutoML} for model compression and acceleration on mobile
  devices.
\newblock In {\em European Conference on Computer Vision (ECCV)}, 2018.

\bibitem{he2017channel}
Yihui He, Xiangyu Zhang, and Jian Sun.
\newblock Channel pruning for accelerating very deep neural networks.
\newblock In {\em International Conference on Computer Vision (ICCV)}, 2017.

\bibitem{hoefler2021sparsity}
Torsten Hoefler, Dan Alistarh, Tal Ben-Nun, Nikoli Dryden, and Alexandra Peste.
\newblock Sparsity in deep learning: Pruning and growth for efficient inference
  and training in neural networks.
\newblock {\em arXiv preprint arXiv:2102.00554}, 2021.

\bibitem{hubara2021accelerated}
Itay Hubara, Brian Chmiel, Moshe Island, Ron Banner, Seffi Naor, and Daniel
  Soudry.
\newblock Accelerated sparse neural training: A provable and efficient method
  to find {N:M} transposable masks.
\newblock In {\em Conference on Neural Information Processing Systems
  (NeurIPS)}, 2021.

\bibitem{hubara2021accurate}
Itay Hubara, Yury Nahshan, Yair Hanani, Ron Banner, and Daniel Soudry.
\newblock Accurate post training quantization with small calibration sets.
\newblock In {\em International Conference on Machine Learning (ICML)}, 2021.

\bibitem{yolov5}
Glenn Jocher.
\newblock {YOLOv5}.
\newblock https://github.com/ultralytics/yolov5, 2022.

\bibitem{kurtic2022optimal}
Eldar Kurtic, Daniel Campos, Tuan Nguyen, Elias Frantar, Mark Kurtz, Benjamin
  Fineran, Michael Goin, and Dan Alistarh.
\newblock The {Optimal BERT Surgeon}: Scalable and accurate second-order
  pruning for large language models.
\newblock {\em arXiv preprint arXiv:2203.07259}, 2022.

\bibitem{pmlr-v119-kurtz20a}
Mark Kurtz, Justin Kopinsky, Rati Gelashvili, Alexander Matveev, John Carr,
  Michael Goin, William Leiserson, Sage Moore, Bill Nell, Nir Shavit, and Dan
  Alistarh.
\newblock Inducing and exploiting activation sparsity for fast inference on
  deep neural networks.
\newblock In {\em International Conference on Machine Learning (ICML)}, 2020.

\bibitem{lecun1990optimal}
Yann LeCun, John~S Denker, and Sara~A Solla.
\newblock Optimal brain damage.
\newblock In {\em Conference on Neural Information Processing Systems
  (NeurIPS)}, 1990.

\bibitem{li2021brecq}
Yuhang Li, Ruihao Gong, Xu~Tan, Yang Yang, Peng Hu, Qi~Zhang, Fengwei Yu, Wei
  Wang, and Shi Gu.
\newblock {BRECQ}: Pushing the limit of post-training quantization by block
  reconstruction.
\newblock In {\em International Conference on Learning Representations (ICLR)},
  2021.

\bibitem{liang2021pruning}
Tailin Liang, John Glossner, Lei Wang, Shaobo Shi, and Xiaotong Zhang.
\newblock Pruning and quantization for deep neural network acceleration: A
  survey.
\newblock {\em Neurocomputing}, 461:370--403, 2021.

\bibitem{lin2014microsoft}
Tsung-Yi Lin, Michael Maire, Serge Belongie, James Hays, Pietro Perona, Deva
  Ramanan, Piotr Doll{\'a}r, and C~Lawrence Zitnick.
\newblock Microsoft {COCO}: Common objects in context.
\newblock In {\em European Conference on Computer Vision (ECCV)}, 2014.

\bibitem{liu2021group}
Liyang Liu, Shilong Zhang, Zhanghui Kuang, Aojun Zhou, Jing-Hao Xue, Xinjiang
  Wang, Yimin Chen, Wenming Yang, Qingmin Liao, and Wayne Zhang.
\newblock Group fisher pruning for practical network compression.
\newblock In {\em International Conference on Machine Learning (ICML)}, 2021.

\bibitem{marcel2010torchvision}
S{\'e}bastien Marcel and Yann Rodriguez.
\newblock Torchvision the machine-vision package of torch.
\newblock In {\em ACM International Conference on Multimedia}, 2010.

\bibitem{2015-martens}
James Martens and Roger Grosse.
\newblock Optimizing neural networks with kronecker-factored approximate
  curvature.
\newblock In {\em International Conference on Machine Learning (ICML)}, 2015.

\bibitem{NVIDIASparse}
Asit Mishra, Jorge~Albericio Latorre, Jeff Pool, Darko Stosic, Dusan Stosic,
  Ganesh Venkatesh, Chong Yu, and Paulius Micikevicius.
\newblock Accelerating sparse deep neural networks.
\newblock {\em arXiv preprint arXiv:2104.08378}, 2021.

\bibitem{nagel2020up}
Markus Nagel, Rana~Ali Amjad, Mart Van~Baalen, Christos Louizos, and Tijmen
  Blankevoort.
\newblock Up or down? {A}daptive rounding for post-training quantization.
\newblock In {\em International Conference on Machine Learning (ICML)}, 2020.

\bibitem{nagel2019data}
Markus Nagel, Mart~van Baalen, Tijmen Blankevoort, and Max Welling.
\newblock Data-free quantization through weight equalization and bias
  correction.
\newblock In {\em International Conference on Computer Vision (ICCV)}, 2019.

\bibitem{nagel2021white}
Markus Nagel, Marios Fournarakis, Rana~Ali Amjad, Yelysei Bondarenko, Mart van
  Baalen, and Tijmen Blankevoort.
\newblock A white paper on neural network quantization.
\newblock {\em arXiv preprint arXiv:2106.08295}, 2021.

\bibitem{nahshan2021loss}
Yury Nahshan, Brian Chmiel, Chaim Baskin, Evgenii Zheltonozhskii, Ron Banner,
  Alex~M Bronstein, and Avi Mendelson.
\newblock Loss aware post-training quantization.
\newblock {\em Machine Learning}, 110(11):3245--3262, 2021.

\bibitem{deepsparse}
NeuralMagic.
\newblock {The DeepSparse Inference Engine}.
\newblock https://github.com/neuralmagic/deepsparse, 2022.

\bibitem{rajpurkar2016squad}
Pranav Rajpurkar, Jian Zhang, Konstantin Lopyrev, and Percy Liang.
\newblock {SQuAD}: 100,000+ questions for machine comprehension of text.
\newblock In {\em Conference on Empirical Methods in Natural Language
  Processing (EMNLP)}, 2016.

\bibitem{reddi2020mlperf}
Vijay~Janapa Reddi, Christine Cheng, David Kanter, Peter Mattson, Guenther
  Schmuelling, Carole-Jean Wu, Brian Anderson, Maximilien Breughe, Mark
  Charlebois, William Chou, et~al.
\newblock Mlperf inference benchmark.
\newblock In {\em 2020 ACM/IEEE 47th Annual International Symposium on Computer
  Architecture (ISCA)}, pages 446--459. IEEE, 2020.

\bibitem{imagenet}
Olga Russakovsky, Jia Deng, Hao Su, Jonathan Krause, Sanjeev Satheesh, Sean Ma,
  Zhiheng Huang, Andrej Karpathy, Aditya Khosla, Michael Bernstein, et~al.
\newblock Imagenet large scale visual recognition challenge.
\newblock {\em International Journal of Computer Vision}, 115(3):211--252,
  2015.

\bibitem{singh2020woodfisher}
Sidak~Pal Singh and Dan Alistarh.
\newblock {WoodFisher}: Efficient second-order approximation for neural network
  compression.
\newblock In {\em Conference on Neural Information Processing Systems
  (NeurIPS)}, 2020.

\bibitem{wang2019eigendamage}
Chaoqi Wang, Roger Grosse, Sanja Fidler, and Guodong Zhang.
\newblock Eigendamage: Structured pruning in the {K}ronecker-factored
  eigenbasis.
\newblock In {\em International Conference on Machine Learning (ICML)}, 2019.

\bibitem{wanghaq}
Kuan Wang, Zhijian Liu, Yujun Lin, Ji~Lin, and Song Han.
\newblock {HAQ}: Hardware-aware automated quantization with mixed precision.
\newblock In {\em Conference on Computer Vision and Pattern Recognition
  (CVPR)}, 2019.

\bibitem{wang2020towards}
Peisong Wang, Qiang Chen, Xiangyu He, and Jian Cheng.
\newblock Towards accurate post-training network quantization via bit-split and
  stitching.
\newblock In {\em International Conference on Machine Learning (ICML)}, 2020.

\bibitem{yang2020automatic}
Haichuan Yang, Shupeng Gui, Yuhao Zhu, and Ji~Liu.
\newblock Automatic neural network compression by sparsity-quantization joint
  learning: A constrained optimization-based approach.
\newblock In {\em Conference on Computer Vision and Pattern Recognition
  (CVPR)}, 2020.

\bibitem{yao2021hawq}
Zhewei Yao, Zhen Dong, Zhangcheng Zheng, Amir Gholami, Jiali Yu, Eric Tan,
  Leyuan Wang, Qijing Huang, Yida Wang, Michael Mahoney, et~al.
\newblock {HAWQ-v3}: Dyadic neural network quantization.
\newblock In {\em International Conference on Machine Learning (ICML)}, 2021.

\bibitem{zhou2021learning}
Aojun Zhou, Yukun Ma, Junnan Zhu, Jianbo Liu, Zhijie Zhang, Kun Yuan, Wenxiu
  Sun, and Hongsheng Li.
\newblock Learning {N:M} fine-grained structured sparse neural networks from
  scratch.
\newblock In {\em International Conference on Learning Representations (ICLR)},
  2021.

\bibitem{zhu2017prune}
Michael Zhu and Suyog Gupta.
\newblock To prune, or not to prune: exploring the efficacy of pruning for
  model compression.
\newblock {\em arXiv preprint arXiv:1710.01878}, 2017.

\end{thebibliography}

\appendix

\tableofcontents
            
\section{Appendix}

\subsection{Proof of Lemma 1 (Row \& Column Removal)}

\begin{proof}
First, we observe that element $j$ in row $i$, i.e. $[\mathbf{A}]_{ij}$, is set to 0 by the equivalent matrix transformation of subtracting $[\mathbf{A}]_{ij}$ times column $i$ denoted by $\mathbf{A}_{:, i}$ divided by the corresponding diagonal element $[\mathbf{A}]_{ii}$ (similarly, elements in column $i$ can be set to 0 by subtracting row $i$). Thus, Lemma \ref{lem:row-element} corresponds to zeroing $\mathbf{H}^{-1}_{pi}$ and $\mathbf{H}^{-1}_{ip}$ for $i \neq p$ via equivalent matrix transformations, or in other words, Gaussian elimination of one row and column.

Next, we apply these equivalent matrix transformations to both sides of the obvious equality $\mathbf{H}^{-1} \mathbf{H} = \mathbf{I}$, which ultimately gives an equation of the following $\mathbf{A} \mathbf{B} = \mathbf{C}$ form:
\begin{equation}
    \begin{bmatrix} 
    	\mathbf{A_1} & \mathbf{0} & \mathbf{A_2} \\
    	\mathbf{0}^\top & a & \mathbf{0}^\top \\
    	\mathbf{A_4} & \mathbf{0} & \mathbf{A_3} \\
	\end{bmatrix}
	\cdot 
	\begin{bmatrix} 
    	\mathbf{B_1} & \mathbf{b_1} & \mathbf{B_2} \\
    	\mathbf{b_4}^\top & b & \mathbf{b_2}^\top \\
    	\mathbf{B_4} & \mathbf{b_3} & \mathbf{B_3} \\
	\end{bmatrix}
	=
	\begin{bmatrix}
	    \mathbf{I} & \mathbf{c_1} & \mathbf{0} \\
    	\mathbf{c_4}^\top & c & \mathbf{c_2}^\top \\
    	\mathbf{0} & \mathbf{c_3} & \mathbf{I} \\
	\end{bmatrix}.
\end{equation}
Notice now that the entries of $\mathbf{B}$ corresponding to the eliminated row and column in $\mathbf{A}$ do not affect the $\mathbf{I}$ and $\mathbf{0}$ blocks in $\mathbf{C}$ since they are always multipled by 0. Thus, the matrix of the $\mathbf{A_i}$ blocks must be the inverse of the $\mathbf{B_i}$ block matrix, which is exactly what we wanted to calculate.
\end{proof}

\subsection{ExactOBS Global Step Pseudocode}

This section provides more details about the global step of the ExactOBS algorithm described in Section \ref{sec:trueobs} in the form of pseudocode.

\begin{algorithm}[H]
    \centering
    \caption{Let $\mathbf{P}$ be a $d_\text{row} \times d_\text{col}$ matrix storing the order in which weights are pruned by ExactOBS in each row and let $\mathbf{L}$ be the matrix of the corresponding loss-changes $\delta \mathcal{L}$. Then the following procedure determines the global OBS mask with $k$ pruned weights.}
    \label{alg:trueobs-row-quant}
    \begin{algorithmic}
        \STATE $Q = \{(i, 0) \, | \, 1 \leq i \leq d_\text{row} \}$
        \FOR {$k$ times}
            \STATE $i, j \gets \text{argmin}_{(i,j) \in Q} \quad [\mathbf{L}]_{i(j + 1)} \,\, \text{if} \,\, j < d_\text{col} \,\, \text{else} \,\, \infty$
            \STATE $Q \gets Q - \{(i, j)\}$
            \STATE $Q \gets Q \cup \{(i, j + 1)\}$
        \ENDFOR
        \STATE $Q$ contains the number of pruned elements $j$ per row $i$, which together with $\mathbf{P}$ yields the mask.
    \end{algorithmic}
    \label{alg:trueobs-global}
\end{algorithm}

For increased efficiency, the set $Q$ can be implemented, for example, as a min-heap. Finally, we note that the slightly simpler method of picking the $k$ smallest elements in $\mathbf{L}$ and then counting how many were picked in each row typically produces essentially the same results as Algorithm~\ref{alg:trueobs-global} in practice since the loss changes generally increase monotonically as more weights are pruned.

\subsection{OBQ-ExactOBS Algorithm Pseudocode}

The OBQ version of the ExactOBS algorithm is given below; we emphasize the similarity to the pruning variant of ExactOBS shown in Algorithm \ref{alg:trueobs-row}.

\begin{algorithm}[H]
    \centering
    \caption{Quantize $k \leq d_\text{col}$ weights from row $\mathbf{w}$ with inverse Hessian $\mathbf{H}^{-1} = (2 \mathbf{X} \mathbf{X}^\top)^{-1}$ according to OBS in $O(k \cdot d_\text{col}^2$) time.}
    \label{alg:trueobs-row-quant}
    \begin{algorithmic}
        \STATE $M = \{1, \dots, d_\text{col}\}$
        \FOR {$i = 1, \dots, k$}
            \STATE $p \gets \text{argmin}_{p \in M} \frac{1}{[\mathbf{H}^{-1}]_{pp}} \cdot (q(w_p) - w_p)^2$
            \STATE $\mathbf{w} \gets \mathbf{w} - \mathbf{H}^{-1}_{:, p} \frac{1}{[\mathbf{H}^{-1}]_{pp}} \cdot (w_p - q(w_p))$
            \STATE $\mathbf{H}^{-1} \gets \mathbf{H}^{-1} - \frac{1}{[\mathbf{H}^{-1}]_{pp}} \mathbf{H}^{-1}_{:, p} \mathbf{H}^{-1}_{p, :}$
            \STATE $M \gets M - \{p\}$
        \ENDFOR
    \end{algorithmic}
\end{algorithm}

\subsection{Further Experiment Details}

We now provide some additional details about our experiments in Section \ref{sec:experiments}.

\paragraph{Bias and Variance Correction.} Although our bias and variance correction step applied to YOLO and BERT models is similar to the schemes described in \cite{nagel2019data} and \cite{banner2019post}, we now describe our exact procedure for additional clarity:

\begin{enumerate}
    \item Sample one batch from the calibration dataset.
    \item Perform inference on this batch with the \textit{dense} model and record after each normalization layer the mean $\mu_\text{dense}^\ell$ and standard deviation $\sigma_\text{dense}^\ell$ for each channel (for CNNs) / feature (for Transformers) over this batch.
    \item Perform inference on this batch with the \textit{compressed} model and record the means $\mu_\text{comp}^\ell$ and standard deviations $\sigma_\text{comp}^\ell$ as in step 2, while already applying mean and variance correction to the layer outputs $X^\ell$ via:
    \begin{equation}
        Y^\ell = \frac{\sigma_\text{dense}^\ell}{\sigma_\text{comp}^\ell} \cdot (X^\ell- \mu_\text{comp}^\ell + \mu_\text{dense}^\ell)
        \label{eq:stat-corr}
    \end{equation}
    \item Merge (\ref{eq:stat-corr}) into the affine parameters of the respective normalization layer.
\end{enumerate}

We note that it is critical to apply the statistics correction already while computing the compressed means and variances in step 3 in order to properly account for compounding distribution shifts.

\paragraph{Non-Uniform Sparsity Choices.} The method we use for determining per-layer (unstructured or blocked) sparsity values to reach a certain overall budget with minimal accuracy loss requires a discrete set of sparsity choices per layer. For both unstructured and blocked sparsity, we follow \cite{frantar2021m} and choose a grid where each point prunes the same fraction of remaining weights $\delta$. Hence, sparsity choice $s_i$ is given by:
\begin{equation}
    s_i = 1 - (1 - \delta)^i.
\end{equation}
In both cases we choose $\delta = 0.9$, which corresponds to pruning 10\% of the remaining weights. For unstructured sparsity, we generate choices until $s_i > 0.99$ and for blocked sparsity until $s_i > 0.95$. We note that these sets of sparsity options are chosen to allow for maximum flexibility. However, in many cases, similar results can likely be achieved with significantly fewer, but more carefully selected (e.g. using the fact that very high sparsities will typically never be chosen for lower FLOP reduction targets), options and thus less required database storage.

\paragraph{Activation Quantization.} In our GPU-focussed quantization + 2:4 pruning experiments we also quantize all activations. This is done by simply optimizing the zero point and quantization scale for one input batch of each layer using exactly the same procedure as for the weights, just on tensor- instead of channel-level (which is the same LAPQ \cite{nahshan2021loss} procedure also used by BRECQ \cite{li2021brecq}). This is again done independently for each layer and the corresponding quantization information is stored in the model database to allow for quick stitching. More advanced schemes such as reoptimizing the weights to better match the quantized inputs (see Appendix \ref{sec:obq-seq}) may be possible, but we found the simple procedure just described to already work quite well.

\subsection{Timing Information}

In this section, we provide detailed information about the runtime of our method. All numbers reported here are for the execution on a single NVIDIA RTX 3090 GPU using our PyTorch implementations. Pruning runs with a global step are performed with the ``less compute'' variant described in Figure \ref{fig:global-trueobs}. Hence an entire database of \textit{many} pruning levels can be generated in approximately the time shown for unstructured and block pruning runs here.

\paragraph{PTQ Runtime Comparison.} We begin with a runtime comparison of existing state-of-the-art post-training methods at the task of quantizating the weights of all layers of a ResNet50 to 4 bits. All timings were collected by executing the authors' open-source implementations on the same hardware, the results are shown in Table \ref{tab:ptq-runtimes}.

\begin{table}[h!]
    \centering
    \begin{tabular}{|l|ccccc|}
        \toprule
        Model & BitSplit & AdaRound & AdaQuant & BRECQ & OBQ \\
        \midrule
        ResNet50 & 124m & 55m & 17m & 53m & 65m \\
        \bottomrule
    \end{tabular}
    \vspace{5pt}
    \caption{Runtimes of post-training quantization methods in minutes (m).}
    \label{tab:ptq-runtimes}
\end{table}

BRECQ, AdaRound and our method OBQ all take around one hour to fully quantize ResNet50, the former two slightly less and the latter slightly more. Meanwhile, BitSplit takes about twice as long, whereas AdaQuant is $3\times$ faster. However, as shown in Table \ref{tab:quant-seq} in the main text (as well as in Table~\ref{tab:quant-indep}), AdaQuant is also considerably less accurate than the other methods. In summary, the runtime of ExactOBS is in line with existing post-training methods. Additional optimizations, like periodically shrinking the Hessian by omitting rows/columns of pruned/quantized weights, can likely improve the practical speed further.

\paragraph{Different Compression Types.} Next, we study the runtime of ExactOBS applied to different types of compression problems. We consider a smaller model (YOLOv5s), a medium model (ResNet50) and a larger one (BERT). The corresponding runtimes for all compression types
featured in this work are listed in Table \ref{tab:trueobs-runtimes}.

\begin{table}[h!]
    \centering
    \begin{tabular}{|l|ccccc|}
        \toprule
        Model & Quant & Unstr & 4-block & 2:4 & Quant 2:4 \\
        \midrule
        ResNet50 & 65m & 64m & 61m & 31m & 35m \\
        YOLOv5s & 7m & 6m & 10m & 3m & 4m \\
        BERT & 111m & 103m & 142m & 51m & 56m \\
        \bottomrule
    \end{tabular}
    \vspace{5pt}
    \caption{Runtimes of ExactOBS for different models and compression types in minutes (m).}
    \label{tab:trueobs-runtimes}
\end{table}

In general, we can see that quantization and unstructured pruning take about the same time, which matches with the fact that the corresponding algorithms are very similar. Correspondingly, 2:4 pruning and quantizing a 2:4 pruned model are only approximately half as expensive, which is again expected as they perform half the work. For YOLO and BERT, blocked pruning is the most expensive compression type due to the overheads incurred by handling the additional $c \times c$ block matrices (see Section \ref{sec:trueobs}). Interestingly, for ResNet50, this is not the case, which is probably related to the highly non-uniform compute distribution that is discussed in more detail in the next paragraph.  Overall, these results show that our techniques are quick for small models and still reasonably efficient even for bigger models like BERT, taking less than 2 hours on a single GPU. Finally, we note that ExactOBS is essentially perfectly parallelizable and its runtime can thus scale linearly with the number of available GPUs.

\begin{figure}[h!]
    \centering
    \includegraphics[width=.66\textwidth]{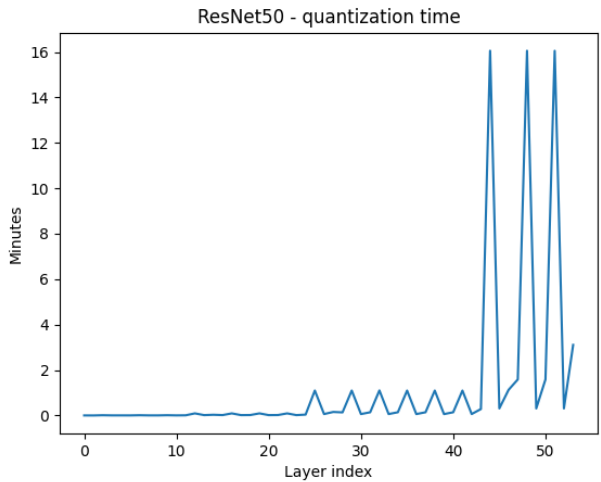}
    \caption{Runtime of OBQ for each layer of ResNet50.}
    \label{fig:rn50-timings}
\end{figure}

\paragraph{Per-Layer Runtimes.} Finally, we note that as the time complexity of OBQ implemented via ExactOBS is $O(d_\text{row} \cdot d_\text{col}^3)$, i.e. cubic in the column dimension, the overall runtime can often be dominated by a few particularly large layers. This is illustrated e.g. by ResNet50 where, as shown in Figure \ref{fig:rn50-timings}, about $75\%$ of the overall runtime is spent in the $3 \times 3$ convolutions of the last block (which have $d_\text{col} \approx 4500$ when unfolded), of which there are just 3 in total. Meanwhile, most of the earlier layers are quantized within seconds. This means that one could, in many cases, reduce the overall compression runtime significantly by applying a faster but less accurate method to just those few bottleneck layers while still achieving more accurate compression on all the others through our techniques.

\subsection{Multiple AdaPrune Iterations}

While AdaPrune \cite{hubara2021accelerated} determined all weights to prune in a single step, the authors of \cite{frantar2021m} found that iterating this process in smaller steps can often improve performance significantly, at quickly increasing computational costs. Our method realizes the very limit of this scheme with one step for each weight. In this section, we study how OBQ comares against AdaPrune with a varying number of pruning and full reoptimization steps. For that purpose, we prune BERT to uniform 75\% sparsity by applying AdaPrune in $k = 2^i$ steps that, as suggested by \cite{frantar2021m}, all prune the same fraction of remaining weights.

\begin{table}[h!]
    \centering
    \begin{tabular}{|l|c|c|ccccc|}
        \toprule
        Model & Sparse & ExactOBS & AP $1\times$ & AP $2\times$ & AP $4\times$ & AP $8\times$ & AP $16\times$ \\
        \midrule
        BERT & 75\% & \textbf{-7.69} & -61.54 & -31.67 & -19.73 & -18.16 & -14.89 \\
        \bottomrule
    \end{tabular}
    \vspace{5pt}
    \caption{Comparing F1 drops against AdaPrune (AP) with a varying number of pruning/reoptimization steps.}
    \label{tab:adaprune-recomp}
\end{table}

Our results confirm the finding of \cite{frantar2021m} that iterating AdaPrune multiple times can significantly improve results, as we see the F1 drop decreasing quickly with just a few such ``recomputations''. Nevertheless, even after 16 full iterations, which have an overall runtime comparable to ExactOBS, the accuracy drop for the (iterative) AdaPrune model is still almost $2\times$ larger than the one of ExactOBS, clearly demonstrating the benefit of our method.

\subsection{Independent Quantization Comparison}

In our uniform quantization experiments in the main paper (see Table \ref{tab:quant-seq}), we only compared OBQ with state-of-the-art \textit{sequential} methods as those are generally significantly more accurate than \textit{independent} ones. However, for completeness, we now additionally compare OBQ with two other methods that have also been used for \textit{independent} layer-wise quantization: BitSplit~\cite{2017-dong} and AdaQuant~\cite{hubara2021accurate}. Here we consider symmetric per-channel quantization as this is the quantization mode BitSplit was designed for. Additionally, we compare ``raw'' quantization performance, that is directly after independent compression without any additional statistics corrections. The results of the comparison are summarized in Table \ref{tab:quant-indep}.

\begin{table}[h!]
    \centering
    \begin{tabular}{|l|ccc|ccc|ccc|}
        \toprule
        \multirow{2}{*}{Method} & \multicolumn{3}{c|}{ResNet18 -- 69.76} & \multicolumn{3}{c|}{ResNet34 -- 73.31} & \multicolumn{3}{c|}{ResNet50 -- 76.13} \\
        & 4bit & 3bit & 2bit & 4bit & 3bit & 2bit & 4bit & 3bit & 2bit \\
        \midrule
        BitSplit & 67.58 & 59.25 & 07.36 & 71.63 & 64.91 & 26.62 & 74.94 & 71.76 & 07.31 \\
        AdaQuant & 65.45 & 49.29 & 00.87 & 69.49 & 56.10 & 00.84 & 72.79 & 53.06 & 00.13 \\
        \midrule
        OBQ (ours) & \textbf{69.18} & \textbf{67.14} & \textbf{48.34} & \textbf{72.85} & \textbf{71.01} & \textbf{51.62} & \textbf{75.50} & \textbf{73.61} & \textbf{46.33} \\
        \bottomrule
    \end{tabular}
    \vspace{5pt}
    \caption{Uniform symmetric per-channel weight quantization.}
    \label{tab:quant-indep}
\end{table}

As expected, OBQ clearly outperforms the other two independent methods on all considered models and bitwidths; at 3 bits by several percent in accuracy and at 2 bits it is the only method that does not break down completely without any statistics correction.

\subsection{Sequential Quantization with OBQ}
\label{sec:obq-seq}

While we primarily focus on the \textit{independent} application of OBC which enables quick stitching of various mixed-compression models, it is also possible to apply OBC sequentially, in similar fashion to state-of-the-art post-training quantization works \cite{nagel2020up, hubara2021accurate, li2021brecq}. While other methods simply perform the per-layer optimization by swapping out the dense model inputs $X_\text{dense}$ for the corresponding inputs in the compressed model $X_\text{comp}$, this does not suffice for OBQ. If the Hessian is computed on $X_\text{comp}$, then the initial dense weights are not a local minimum (with 0 gradient) anymore, hence violating a key assumption of OBQ. Fortunately, this problem can be easily resolved by reoptimizing the dense weights for the new inputs via the closed form solution of linear regression $\mathbf{W}^\top = (\mathbf{X}\mathbf{X}^\top)^{-1}\mathbf{X}\mathbf{Y}^\top$, after which the gradient is 0 again, and OBQ can be applied correctly. We note that $\mathbf{X}\mathbf{Y}^\top$ is a $d_\text{col} \times d_\text{row}$ matrix which can be easily accumulated over multiple batches similar to the OBQ Hessian $2\mathbf{X}\mathbf{X}^\top$, without any major increase in memory consumption.

As a demonstration, we apply sequential OBQ to the task of quantizating ResNet18 to various bitwidths (in the same setup as in Table \ref{tab:quant-seq} in the main paper) and report the results in Table \ref{tab:seq-obq}. Interestingly, for 4 and 3 bits, the results are essentially the same as for the independent version (with batchnorm statistics correction); only for the 2 bits setting there seems to be a noticeable benefit, catching up with the corresponding BRECQ result. A more detailed investigation of this phenomenon could be a interesting direction for future work.

\begin{table}[ht!]
    \centering
    \begin{tabular}{|l|ccc|}
        \toprule
        \multirow{2}{*}{Method} & \multicolumn{3}{c|}{ResNet18 -- 69.76} \\
        & 4bit & 3bit & 2bit \\
        \midrule
        AdaRound & 69.34 & 68.37 & 63.37 \\
        AdaQuant & 68.12 & 59.21 & 00.10 \\
        BRECQ & 69.37 & 68.47 & 64.70 \\
        \midrule
        OBQ + BNT & 69.56 & 68.69 & 64.04 \\
        OBQ -- sequential & 69.56 & 68.68 & 64.93 \\
        \bottomrule
    \end{tabular}
    \vspace{5pt}
    \caption{Comparison with sequential OBQ.}
    \label{tab:seq-obq}
\end{table}

\subsection{Impact of ImageNet Data Augmentations}

As described in the main submission text, for ImageNet experiments, we expand our calibration set with standard data augmentations by a factor of $10$. The is mainly done to ensure that the $2048 \times 2048$ Hessian corresponding to the fully-connected layer of ResNet50 is full rank (which is not the case for just $1024$ images) and thus avoid any hyper-parameter tuning of a dampening constant. Additionally, it should serve as a demonstration that augmentations are cheap to use in conjunction with our method, which is not the case for other post-training methods that would require either considerably increased memory (storing many more activations) or runtime (performing full inference on the entire model for each batch in the per-layer optimization).

We now study the impact of these augmentations on our results, for which rerun OBQ (in the setup of Table \ref{tab:quant-seq} without them, but using dampening $\lambda = 1$ (relative to the values in the Hessian this is actually a rather small constant) for the last layer of ResNet50. A comparison with the original results is shown in Table \ref{tab:augmentations}.

\begin{table}[ht!]
    \centering
    \begin{tabular}{|l|ccc|ccc|}
        \toprule
        \multirow{2}{*}{Method} & \multicolumn{3}{c|}{ResNet18 -- 69.76} & \multicolumn{3}{c|}{ResNet50 -- 76.13} \\
        & 4bit & 3bit & 2bit & 4bit & 3bit & 2bit \\
        \midrule
        OBQ & 69.56 & 68.69 & 64.04 & 75.72 & 75.24 & 70.71 \\
        OBQ -- no aug & 69.59 & 68.51 & 63.87 & 75.87 & 75.06 & 70.51 \\
        \bottomrule
    \end{tabular}
    \vspace{5pt}
    \caption{The impact of data augmentations.}
    \label{tab:augmentations}
\end{table}

As can be seen, the difference between using and not using data augmentations is generally only rather minor at $\approx 0.1 - 0.2\%$. Nevertheless, augmentations are very cheap to use in conjunction with our methods (they only need to be accumulated into the initial per-layer Hessians once) and at the same time avoid a dampening hyper-parameter in several cases; therefore we use them in our ImageNet experiments.

\subsection{Sensitivity to Random Seeds}

For a fixed calibration dataset, the ExactOBS algorithm is deterministic. For ResNet models, small amounts of additional randomness are added by the data augmentations that are applied to the calibration dataset as well as by batchnorm tuning which happens with randomly sampled batches; for the other models we consider, there is no extra randomness beyond the initial sampling of the calibration dataset. To assess how much the results of our methods are affected by these random factors, we quantize ResNet18 to 4bit (symmetric per-channel) and prune ResNet50 to the 2:4 pattern, for 5 different seeds each, and report mean standard deviation in Table \ref{tab:randomness}.

\begin{table}[h!]
    \centering
    \begin{tabular}{|c|c|}
        \toprule
        ResNet18 -- 4bit & ResNet50 -- 2:4 \\
        \midrule
        $69.28 \pm 0.07$ & $74.74 \pm 0.05$\\
        \bottomrule
    \end{tabular}
    \vspace{5pt}
    \caption{Typical random variation of OBC results.}
    \label{tab:randomness}
\end{table}

In conclusion, the variation of results with respect to random seeds is generally very low, here $< 0.1\%$, which is in line with other post training methods \cite{nagel2020up, li2021brecq}.

\subsection{Compound Compression Comparisons}

\begin{figure}[h!]
    \centering
    \begin{subfigure}{.32\textwidth}
      \centering
      \includegraphics[width=\linewidth]{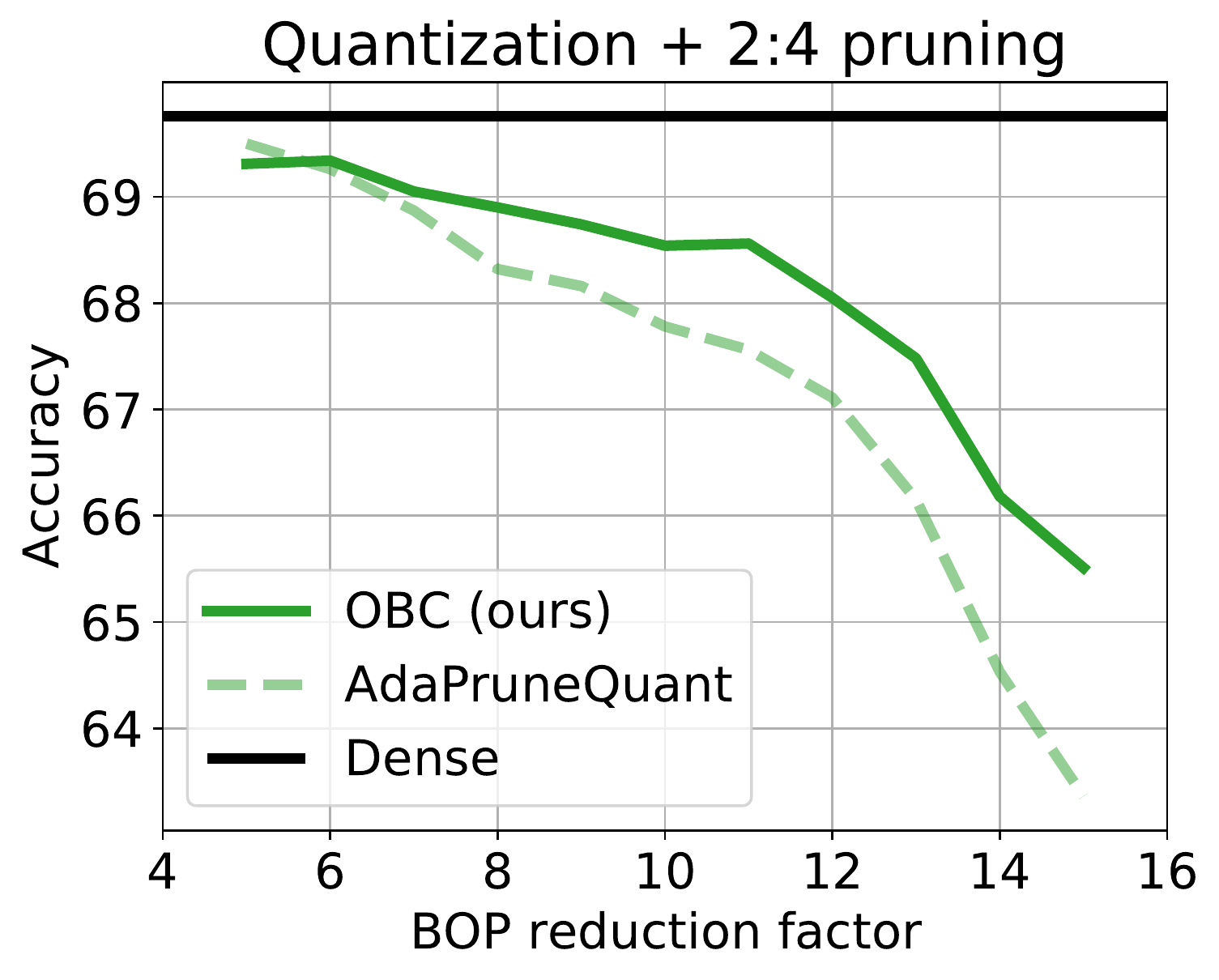}
      \caption{ResNet18.}
    \end{subfigure}
    \begin{subfigure}{.32\textwidth}
      \centering
      \includegraphics[width=\linewidth]{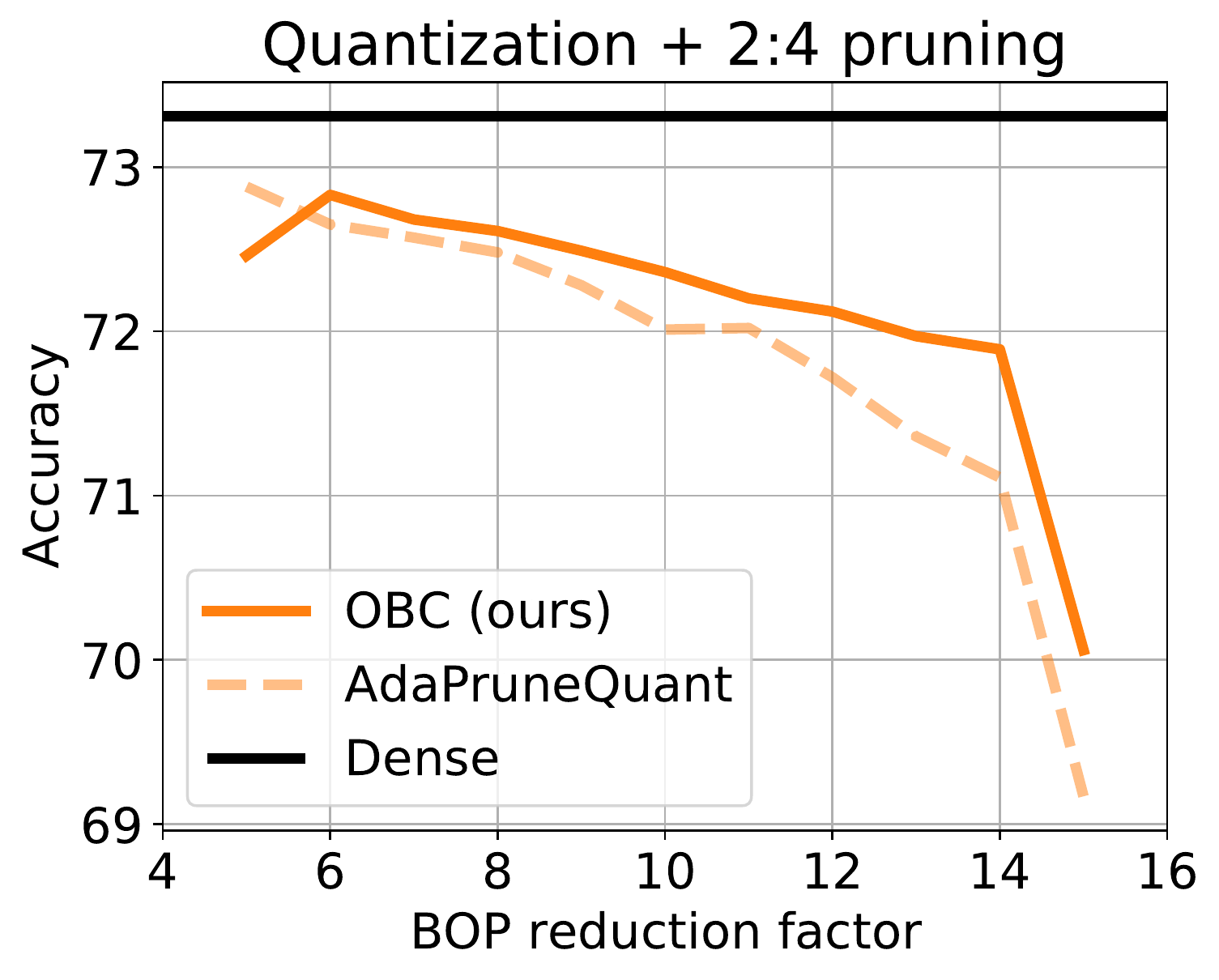}
      \caption{ResNet34.}
    \end{subfigure}
    \begin{subfigure}{.32\textwidth}
      \centering
      \includegraphics[width=\linewidth]{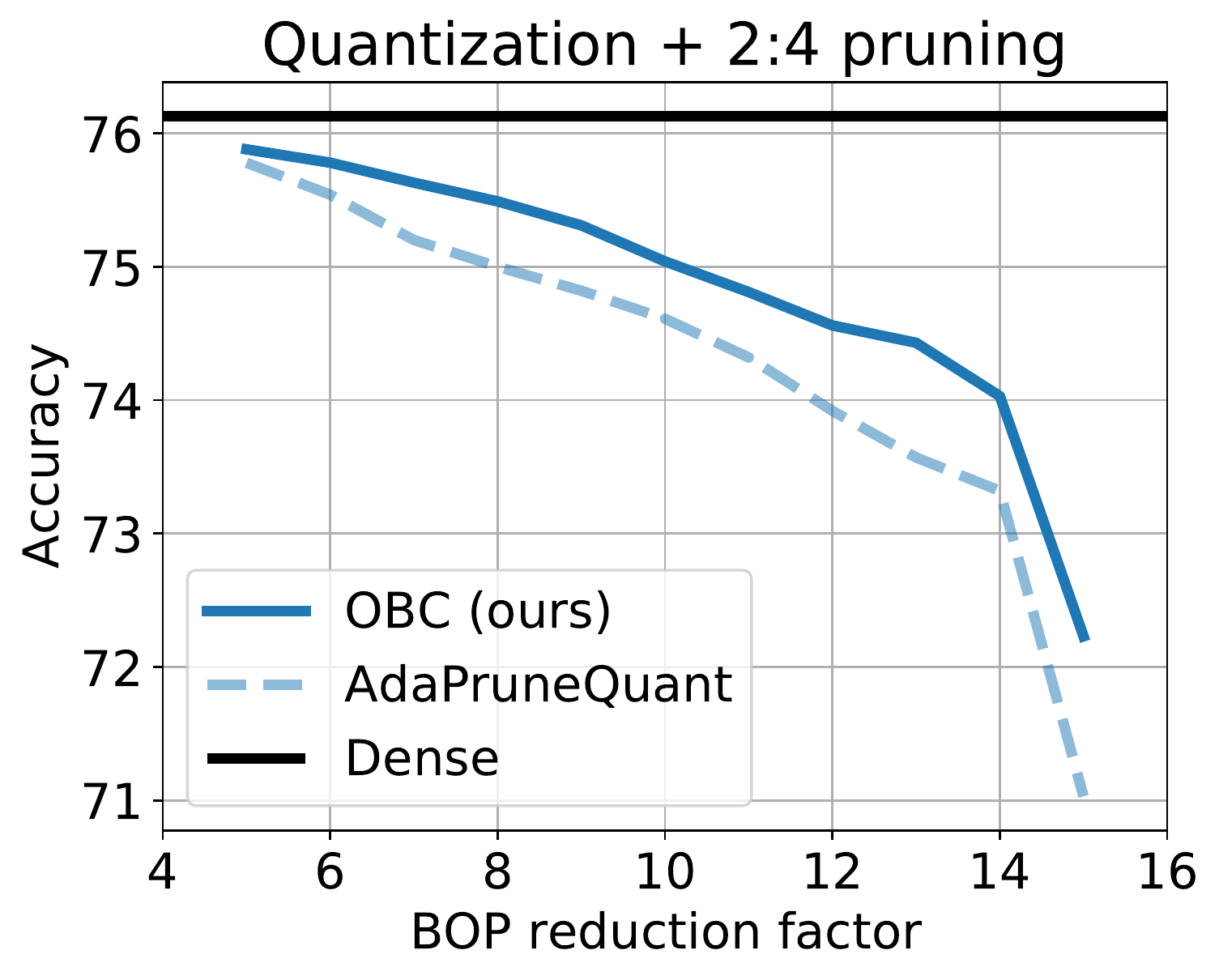}
      \caption{ResNet50.}
    \end{subfigure}
    \caption{Mixed quantization and 2:4 pruning for various BOP reduction targets on ResNet models.}
    \label{fig:mixed-comp-rn}
\end{figure}

\begin{figure}[h!]
    \centering
    \begin{subfigure}{.32\textwidth}
      \centering
      \includegraphics[width=\linewidth]{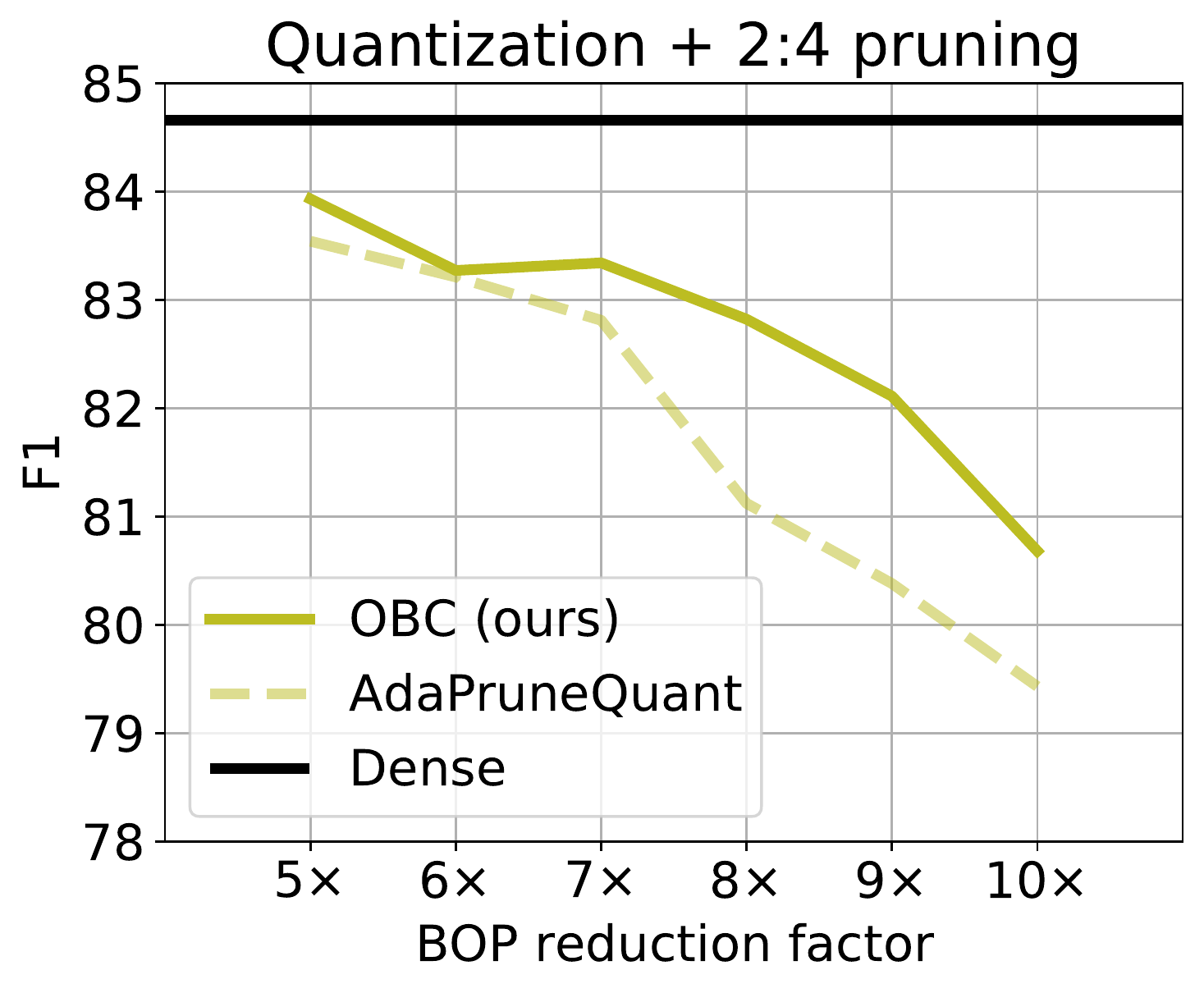}
      \caption{BERT3.}
    \end{subfigure}
    \begin{subfigure}{.32\textwidth}
      \centering
      \includegraphics[width=\linewidth]{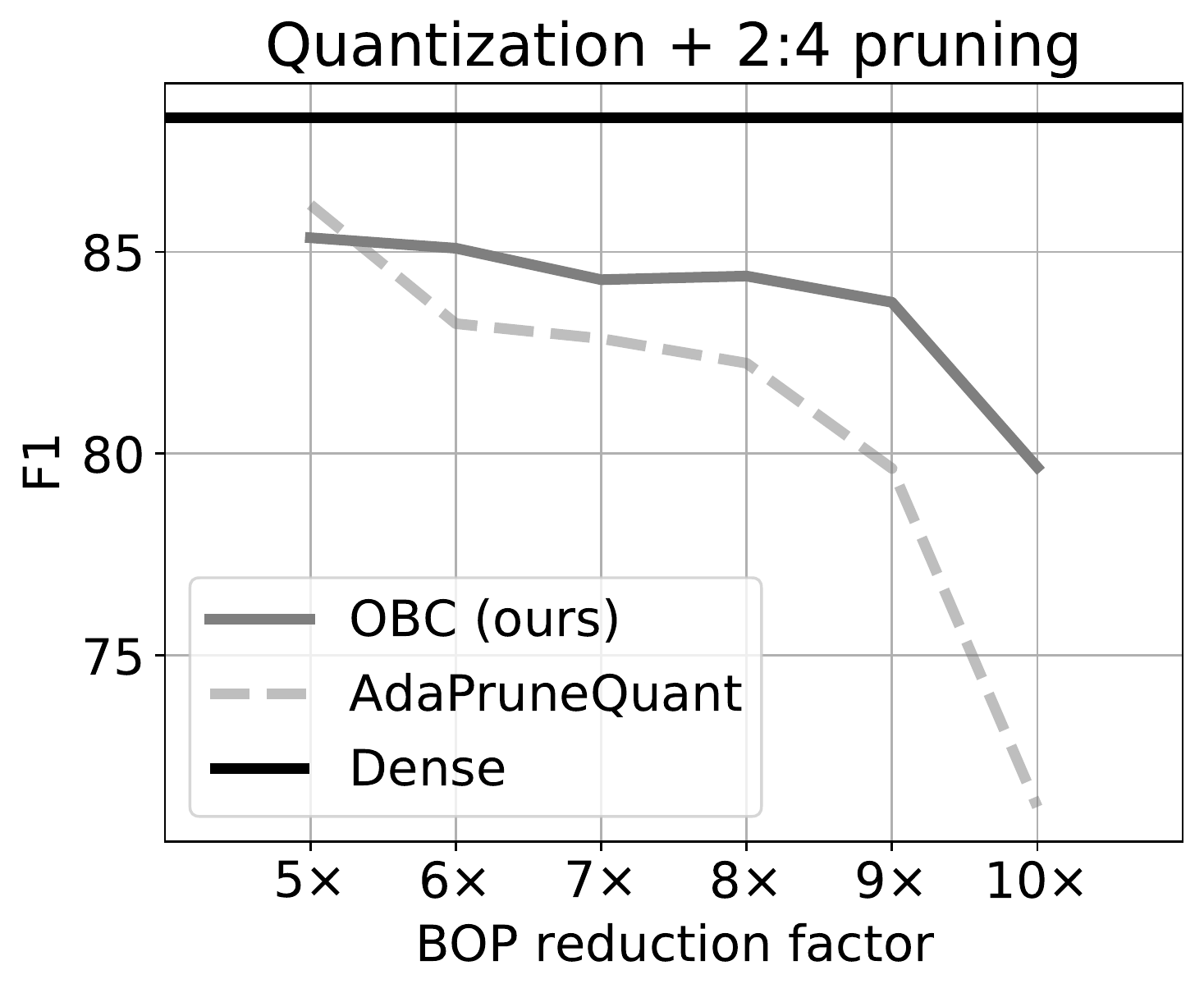}
      \caption{BERT6.}
    \end{subfigure}
    \begin{subfigure}{.32\textwidth}
      \centering
      \includegraphics[width=\linewidth]{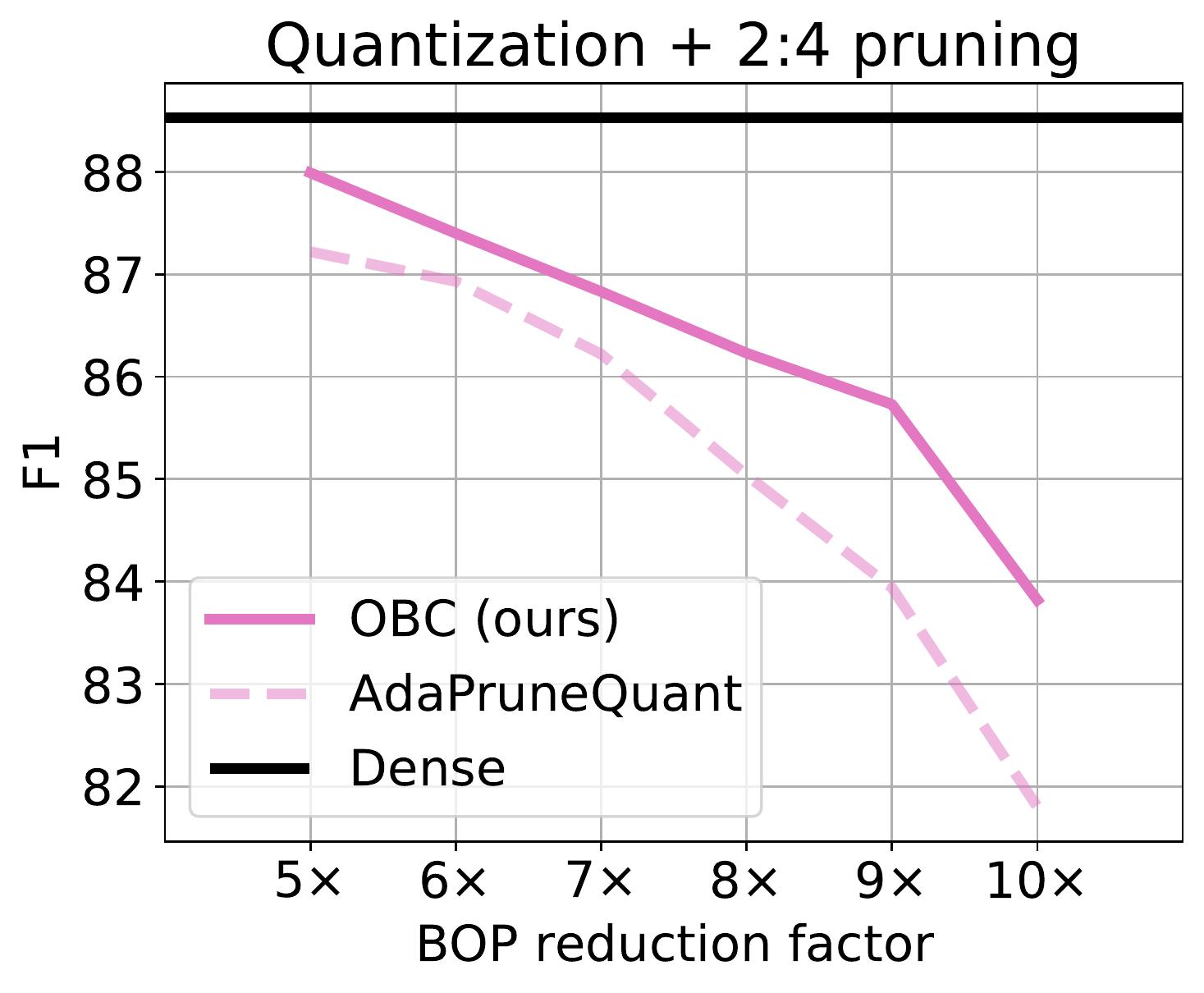}
      \caption{BERT.}
    \end{subfigure}
    \caption{Mixed quantization and 2:4 pruning for various BOP reduction targets on BERT models.}
    \label{fig:mixed-comp-bert}
\end{figure}

\begin{figure}[h!]
    \centering
    \begin{subfigure}{.32\textwidth}
      \centering
      \includegraphics[width=\linewidth]{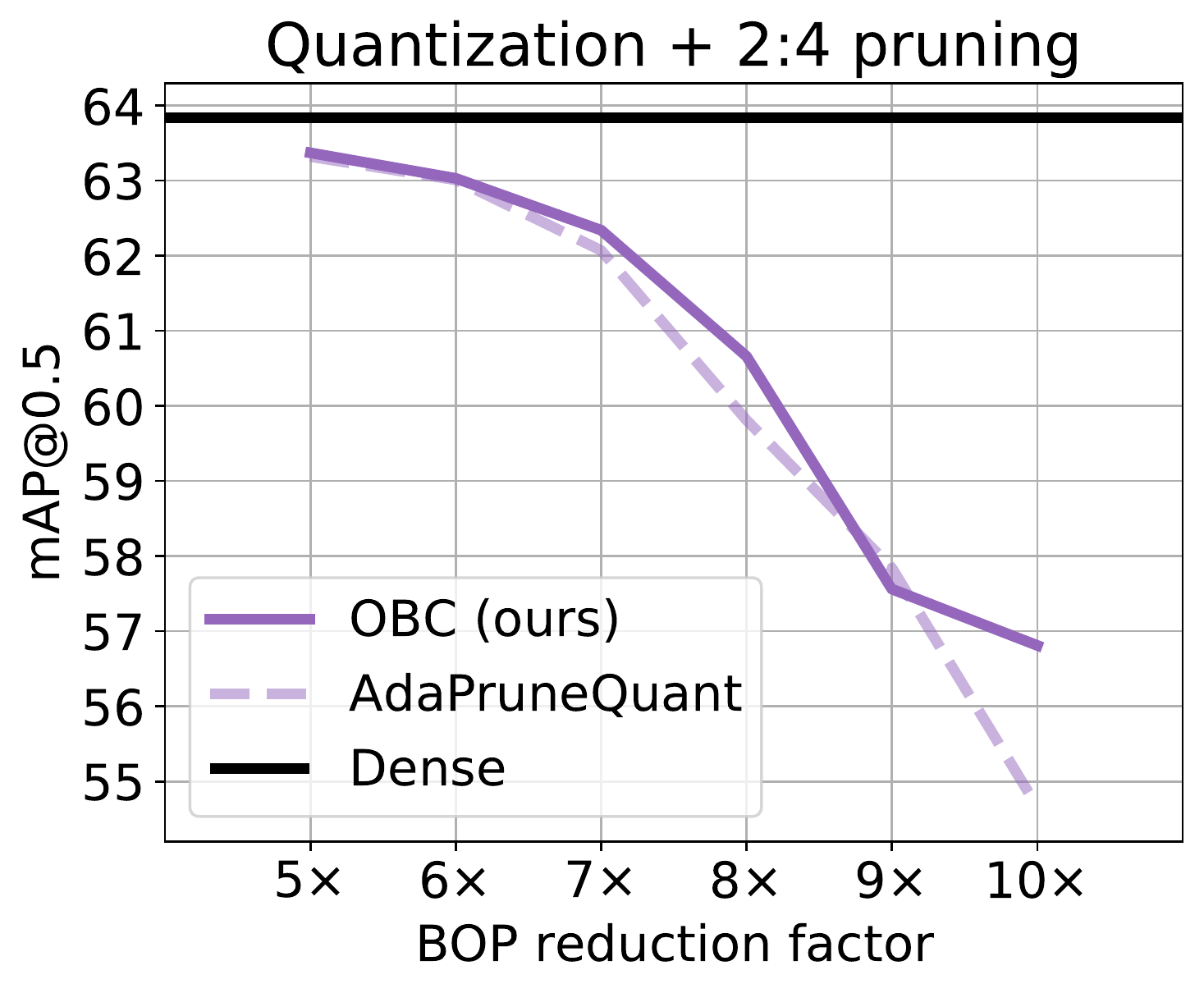}
      \caption{YOLOv5m.}
    \end{subfigure}
    \begin{subfigure}{.32\textwidth}
      \centering
      \includegraphics[width=\linewidth]{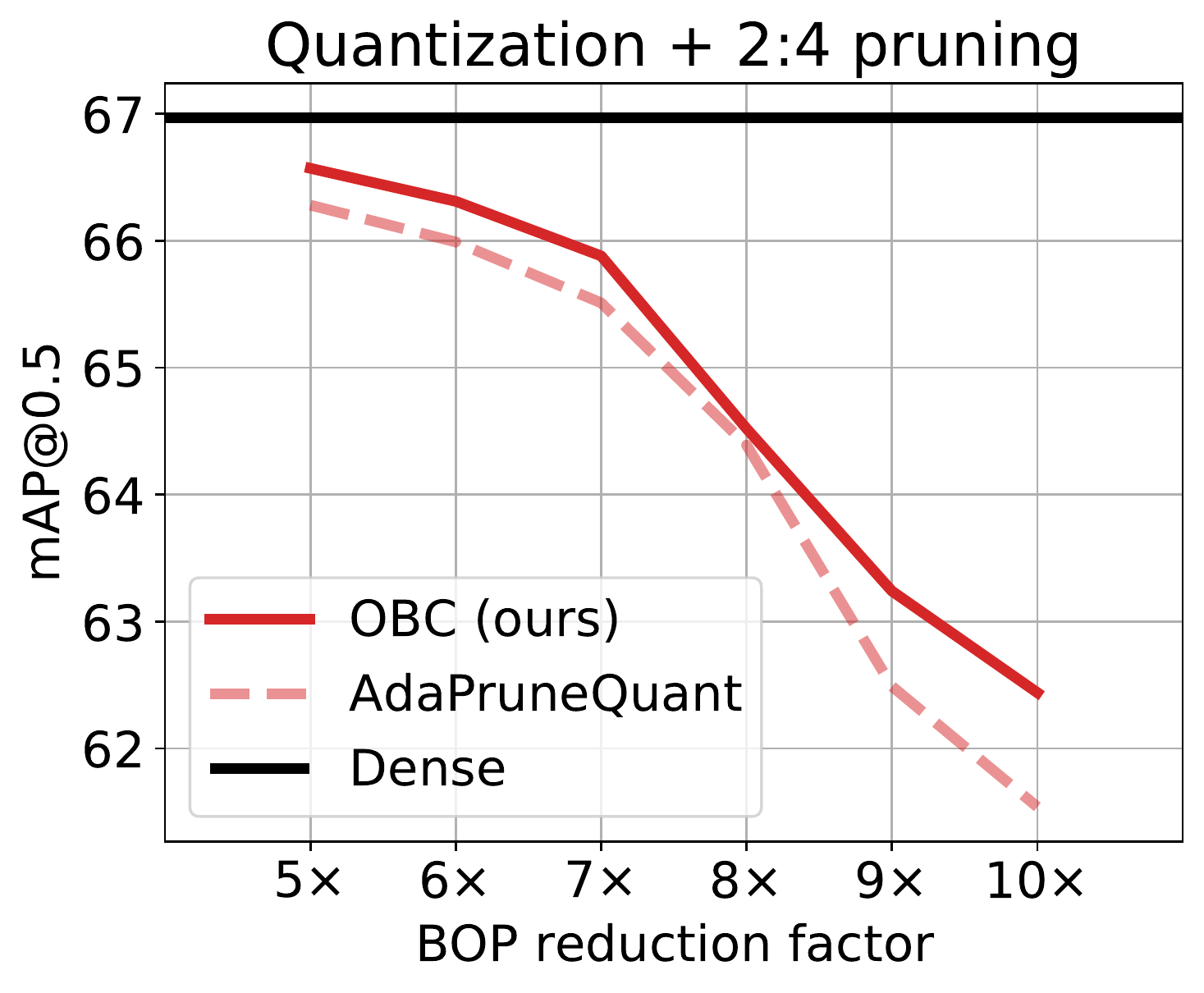}
      \caption{YOLOv5l.}
    \end{subfigure}
    \begin{subfigure}{.32\textwidth}
      \includegraphics[width=\linewidth]{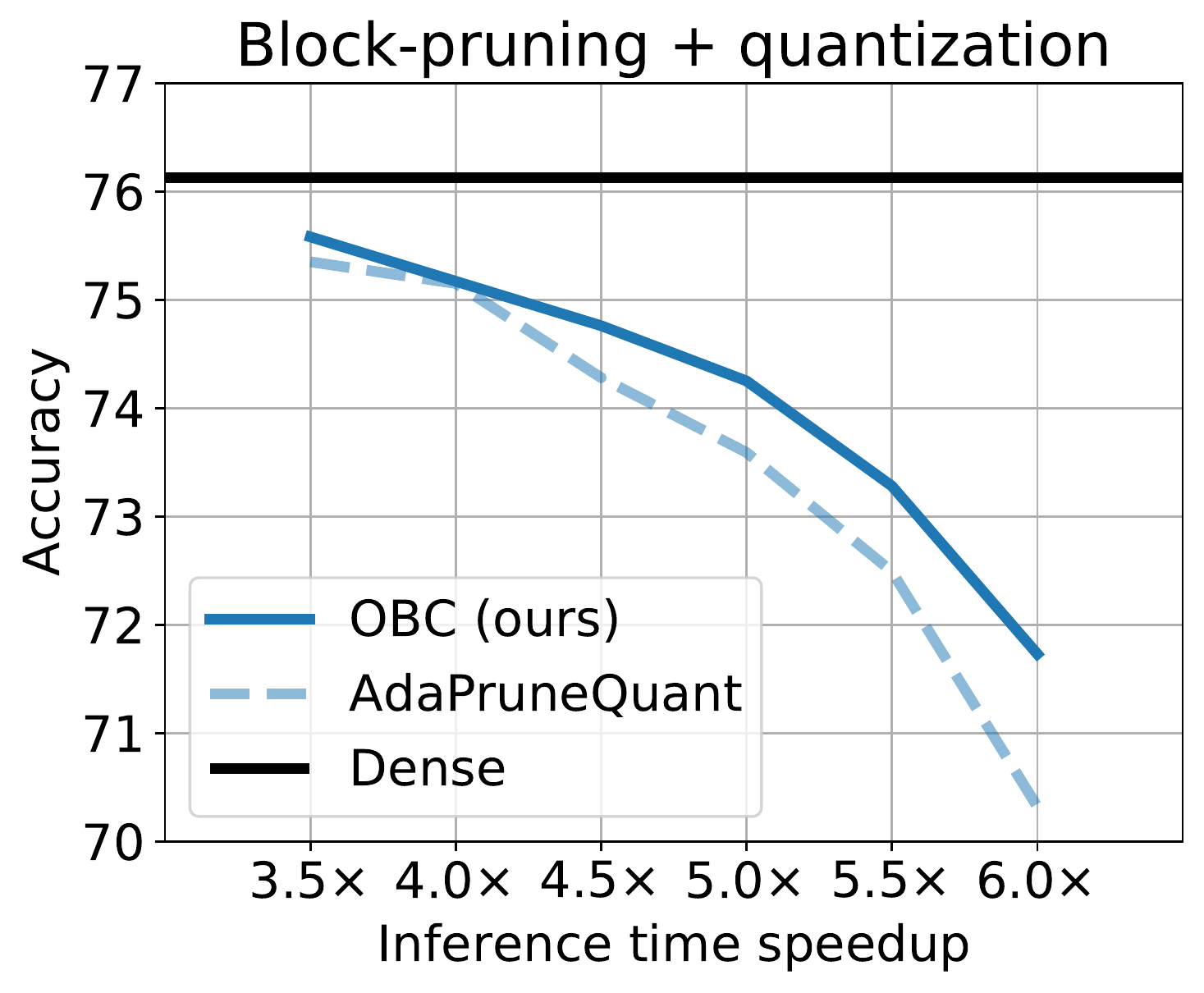}
      \caption{ResNet50 -- CPU.}
    \end{subfigure}
    \caption{(a) \& (b): Mixed quantization and 2:4 pruning for various BOP reduction targets on YOLO models. (c) Block sparsity \& quantization for real-time CPU inference speedup targets on ResNet50.}
    \label{fig:mixed-comp-yolo}
\end{figure}

In the main paper, we focused on independent comparisons for quantization and pruning since existing methods are generally only designed for a single compression approach. In this section, we additionally provide compound comparisons for our GPU and CPU scenarios which combine sparsity and quantization. In particular, we construct a strong baseline by substituting OBC in our mixed setup with the best independent layer-wise pruning and quantization methods, AdaPrune and AdaQuant, respectively. We now provide detailed comparisons for all experiments of Figure \ref{fig:mixed} from the main text, in Figures \ref{fig:mixed-comp-rn}, \ref{fig:mixed-comp-bert} and \ref{fig:mixed-comp-yolo}.

In summary, it appears that, as expected, the accuracy improvements for the individual compression types shown by the experiments in Section \ref{sec:experiments} also transfer to the combined setting. More concretely, for the reduction target ranges highlighted in the main paper, that is $12-14\times$ for ResNet models and $7-8\times$ for others, there is a consistent $0.5-1.5$ point gap between OBC and the AdaPruneQuant baseline. For lower BOP reduction / inference time speedup targets, the gap is typically smaller, which is expected as only the less sensitive layers have to compressed more than to the generally very easy 8-bit level. In contrast, the gaps are largest for the highest targets that also require high compression of sensitive layers as this is where the effects of OBC's more accurate layer-wise compression become particularly noticeable.

\vspace{-10pt}
\vspace{-10pt}

\end{document}